%% file: custom.tex
\pdfoutput=1
\documentclass[11pt]{article}

\usepackage[final]{acl}
\usepackage{times}
\usepackage{latexsym}

\usepackage[T1]{fontenc}
\usepackage[utf8]{inputenc}

\usepackage{microtype}

\usepackage{inconsolata}

\usepackage{inconsolata}
\usepackage{graphicx}
\usepackage{multirow}
\usepackage{subcaption}
\usepackage{caption}
\usepackage{booktabs}
\usepackage{tcolorbox}
\usepackage{enumitem}
\usepackage{amsmath}
\usepackage{makecell}
\usepackage{graphicx}
\usepackage{array}
\usepackage{amsmath}

\title{Large Language Models Are Cross-Lingual Knowledge-Free Reasoners}

\usepackage{pifont}

\author{
Peng Hu\textsuperscript{$\clubsuit$}\thanks{*Equal contribution}, Sizhe Liu\textsuperscript{$\clubsuit$}\footnotemark[1], Changjiang Gao\textsuperscript{$\clubsuit$}, 
\textbf{Xin Huang}\textsuperscript{$\diamond$}, \textbf{Xue Han}\textsuperscript{$\diamond$}, \\ \textbf{Junlan Feng}\textsuperscript{$\diamond$}, \textbf{Chao Deng}\textsuperscript{$\diamond$}, \textbf{Shujian Huang}\textsuperscript{$\clubsuit$} \\
\textsuperscript{$\clubsuit$}National Key Laboratory for Novel Software Technology, Nanjing University \\
\textsuperscript{$\diamond$}China Mobile Research, Beijing, China \\
\texttt{\{hup, liusz, gaocj\}@smail.nju.edu.cn, huangsj@nju.edu.cn} \\
\texttt{\{huangxinyjy, hanxueai, fengjunlan, dengchao\}@chinamobile.com}
}

\begin{document}
\maketitle
\begin{abstract}
Large Language Models have demonstrated impressive reasoning capabilities across multiple languages. However, the relationship between capabilities in different languages is less explored. In this work, we decompose the process of reasoning tasks into two separated components: knowledge retrieval and knowledge-free reasoning, and analyze the relationship between cross-lingual transferability and these two components. With adapted commonsense reasoning datasets and constructed knowledge-free reasoning datasets, we show that the knowledge-free reasoning capability can be nearly perfectly transferred across various source-target language directions despite the secondary impact of resource in some specific target languages, while cross-lingual knowledge retrieval significantly hinders the transfer. Moreover, by analyzing the hidden states and feed-forward network neuron activation during the reasoning, we show that higher similarity of hidden representations and larger overlap of activated neurons could explain the better cross-lingual transferability of knowledge-free reasoning than knowledge retrieval. Thus, we hypothesize that knowledge-free reasoning shares similar neurons in different languages for reasoning, while knowledge is stored separately in different languages.

\end{abstract}

\section{Introduction}
Large language models (LLMs) today have shown strong multitask and multilingual performance in various domains \cite{huang2022towards}, including robust reasoning capabilities across multiple languages \cite{ranaldi-etal-2024-tree}, even for low-resource languages in the training corpus \cite{shi2022language}. 

Previous study reveals that these multilingual LLMs possess certain ability of multilingual transfer \cite{qi2023crosslingual,gao2024multilingual,ye2023language}, which means the skills or knowledge learned with one language can be automatically transferred to another language without extra training. 
% This ability is also referred to as cross-lingual alignment \cite{gao2024multilingual} or cross-lingual consistency \cite{}. 
However, the effect of such cross-lingual transfer varies across tasks. In certain tasks, especially knowledge retrieval tasks, current LLMs show unsatisfactory cross-lingual transfer \cite{qi2023crosslingual,gao2024multilingual}, while in certain reasoning tasks, more effective transfer is observed \cite{ye2023language}. Previous study lacks the analysis on the difference between these tasks, and does not dig further into the specific factors affecting the transfer effectiveness.

\begin{figure}[th]
% \vspace{-0.2cm}
    \centering
    \includegraphics[width=0.49\textwidth]{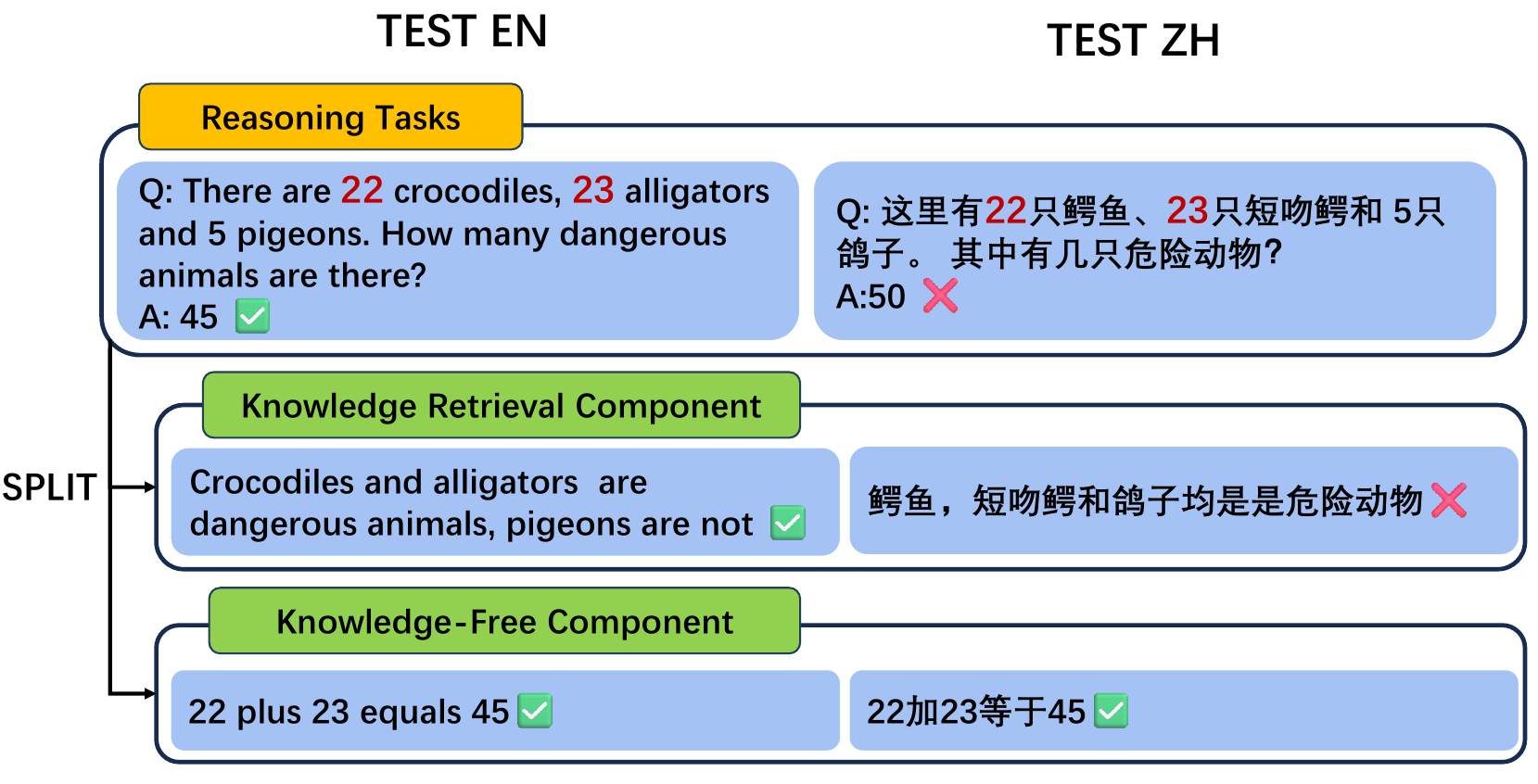}
    \caption{Cross-lingual transfer involves training a model in one language and evaluating it in another. In this figure, the scenario depicts training in English. Reasoning tasks encompass both knowledge retrieval and knowledge-free reasoning. The cross-lingual transfer ratio is significantly lower for knowledge retrieval tasks (e.g., ZH case in EN: "Crocodiles, alligators, and pigeons are dangerous animals") compared to knowledge-free reasoning tasks, which transfer well across languages (e.g., ZH case in EN: "22 plus 23 equals 45").}
    \label{fig:reasoning-type}
% \vspace{-0.2cm}
\end{figure}

% In this study, we divide a reasoning task into two separated components: knowledge retrieval and knowledge-free reasoning, and investigate the transfer effectiveness of them in current LLMs.
% 直观上来说，抽象的推理是一种语言无关的能力，其应该可以很好在不同语言直接传递。general reasoning task在不同语言之间的性能差异可能是由于一些其他因素导致。
Intuitively, abstract reasoning is considered a language-agnostic ability, and thus it should transfer well across languages. The observed performance differences in general reasoning tasks across languages may arise from other factors. 
In this study, we divide a general reasoning task into two separated components: knowledge retrieval and knowledge-free reasoning.
The former means recalling factual knowledge from pre-training~\footnote{Similar to closed-book QA tasks},
% The former means recalling certain information from pre-training, ~\citep{}
while the latter refers to organizing the given knowledge in the context to perform inference and derive a final answer\footnote{Such as Coin Flip~\cite{wei2022chain}}.
Figure~\ref{fig:reasoning-type} provides a clearer understanding of these two components and illustrates the cross-lingual transfer issues explored in this paper.
% \footnote{Based on such definition, we can divide reasoning tasks into three categories: knowledge retrieval tasks that only involve knowledge retrieval without knowledge-free reasoning, knowledge-free reasoning tasks that do not involve knowledge retrieval at all, and common reasoning tasks that require both knowledge retrieval and knowledge-free reasoning.}
% Based on such definition, we can divide reasoning tasks into three categories: knowledge retrieval tasks that only involve knowledge retrieval without knowledge-free reasoning, knowledge-free reasoning tasks that do not involve knowledge retrieval at all, and common reasoning tasks that require both knowledge retrieval and knowledge-free reasoning.
% This distinction is illustrated in Figure~\ref{fig:reasoning-type}.
%mostly in the form of a simple QA~\citep{clark2019boolq,min2020ambigqa,longpre2021mkqa}, 
%covering most existing reasoning tasks, 
%with rarely existing datasets fitting in this category.
% , for which kind of task we will purpose a dataset in this paper.
% In this paper, we focus on cross-lingual transfer on the second and third categories of reasoning tasks, as they are less explored in the previous work.
% We perform evaluation and interpretability analysis on LLaMA-2 models \cite{touvron2023llama} to compare the cross-lingual transferability of retrieval and reasoning ability.

This paper includes both an evaluation part and an interpretability analysis part.
In the evaluation part, we focus on the impact of knowledge retrieval component on cross-lingual transfer in reasoning tasks, and the transferability of knowledge-free reasoning capability, by adapting existing reasoning datasets with different levels of knowledge retrieval demand and creating a clean knowledge-free reasoning dataset, which only includes the knowledge-free reasoning component.
% In the evaluation part, by controlling the requirement of knowledge retrieval in the StrategyQA dataset and constructing a knowledge-free reasoning dataset, we compare the cross-lingual transfer effectiveness of knowledge-involved and knowledge-free reasoning.This paper includes both an evaluation part and an interpretability analysis part.
In the interpretability analysis part, we assess the cross-lingual computational similarity of hidden states and Feed-Forward Network neuron activation to trace and compare the computational process of cross-lingual transfer of knowledge retrieval and knowledge-free reasoning components in LLMs.
Our main findings are:
\begin{itemize}
% \vspace{-0.1cm}
    \item Retrieval component significantly hinders cross-lingual transfer of reasoning tasks. The more knowledge retrieval is required in the task, the lower effectiveness of cross-lingual transfer is observed.

% \vspace{-0.1cm}
    
    \item The ability of knowledge-free reasoning component can be near-perfectly transferred to other languages after fine-tuning in one, while the model's language proficiency in the target languages is also important. 
    %.
    
% \vspace{-0.1cm}
    
    \item The overall cross-lingual computational similarity for knowledge-free reasoning tasks is significantly higher than for knowledge retrieval tasks, especially in the middle-high layers, which are primarily used for reasoning \citep{zhao2024large,wendler2024llamas}. This suggests a language-shared reasoning mechanism in multilingual LLMs.
    % The overall cross-lingual computational similarity of knowledge-free reasoning tasks is significantly higher than knowledge retrieval tasks. Also, the difference in similarity is the most significant in the middle-high layers of the model, which some studies\citep{zhao2024large,wendler2024llamas} suggest are primarily used for reasoning. This suggests a language-shared mechanism of reasoning ability in multilingual LLMs. 

% \vspace{-0.1cm}
    
\end{itemize}

\section{Evaluation Methodology}
\subsection{Overview}
Our evaluation focuses on two main aspects:
% \vspace{-0.1cm}
\paragraph{Impact of Knowledge Retrieval Demand on Cross-Lingual Transfer in Reasoning Tasks}
We aim to analyze how varying levels of knowledge retrieval demand affect cross-lingual transfer in reasoning tasks.
For this purpose, we leverage the commonsense reasoning datasets that provide questions along with several facts required to answer them.
By controlling the number of facts provided to the model within the context, we can manipulate the levels of demand for knowledge retrieval.
As more facts are provided, the model relies less on its internal knowledge storage.
This controlled setup enables us to analyze how the demand for knowledge retrieval influences the cross-lingual transfer of the overall reasoning abilities.

% \vspace{-0.1cm}
\paragraph{Cross-Lingual Transfer of Knowledge-Free Reasoning}
We also aim to investigate the cross-lingual transfer of knowledge-free reasoning, which is less explored in previous work. However, existing reasoning datasets often contain some degree of knowledge retrieval. For instance, while StrategyQA~\citep{geva2021strategyqa} provides knowledge required for reasoning, it is not always complete. Similarly, certain mathematical datasets, like ASDiv, require knowledge retrieval for some problems (as demonstrated in Appendix \ref{sec:app-why-create}). This introduces noise when evaluating the cross-lingual transfer of knowledge-free reasoning. To address this, we constructed a new dataset, the Knowledge-Free Reasoning Dataset (KFRD), which entirely eliminates the need for knowledge retrieval. In addition, we selected several existing datasets that, to the best extent possible, meet the requirements of knowledge-free reasoning to further validate our conclusions. A more detailed explanation of why we constructed KFRD and the dataset selection criteria can be found in Appendix \ref{sec:app-why-create}.

% 在evaluation部分，我们主要想研究的有两点，
% 第一是cross-lingual transfer情况随着Knowledge Retrieval需求的变化。我们使用StrategyQA和QASC数据集进行此研究，这些数据集为常识推理数据集，其在提供问题与回答的同时，还标注了多条回答问题所需的知识。我们可以控制给定模型的知识数量，来控制Knowledge Retrieval的需求。提供的知识越多，模型推理时所需要进行Knowledge Retrieval越少，越接近Knowledge-Free Reasoning.
% 第二是Knowledge-Free Reasoning的cross-lingual transfer情况 我们希望对Knowledge-Free Reasoning的跨语言迁移能力进行研究，但现有的推理数据集或多或少都会包含Knowledge Retrieval的部分，例如StrategyQA虽然给出了推理所需的知识，但是并不完全。一些数学数据集(如ASDiv)的部分题目也需要知识检索。这会对测出的Knowledge-Free Reasoning跨语言传递能力造成干扰。为此，我们新构造了一个纯粹不含知识检索的数据集KFRD（Knowledge-Free Reasoning Dataset）进行分析研究。除此之外，我们选择了一些尽可能满足Knowledge-Free Reasoning的现存数据集来验证我们的结论。
\subsection{Datasets}
This section introduces the datasets used for evaluation. More details on the datasets and the construction process are in Appendix \ref{sec:app-dataset-detail}.
\subsubsection{Reasoning dataset with variable knowledge retrieval demand}

We adapt the popular commonsense reasoning datasets, StrategyQA and QASC~\citep{khot2020qasc}, to analyze the impact of knowledge retrieval on cross-lingual transfer.
They provide pieces of evidence from Wikipedia for answering the question.
Examples can be found in Table~\ref{tab:sqaExample}.% in Appendix~\ref{sec:app-exp-details}.

Namely, we design two kinds of scenarios with variable knowledge retrieval demand in the experiments:
\begin{itemize}

% \vspace{-0.1cm}

\item No Fact (NF): The model is given only the questions. %  and must retrieve the necessary knowledge from itself

% \vspace{-0.1cm}

\item With Fact (WF): The model is provided with the questions and some of the evidence. To control the degree of knowledge retrieval needed, we further devide the WF-1, WF-2 and WF-all settings, where one piece, two pieces, and all pieces of evidence is provided for each question.

% \vspace{-0.1cm}

\end{itemize}

\subsubsection{Knowledge-free reasoning dataset}

% To rule out such effect and focus on the knowledge-free reasoning process, we construct a data that minimizes the need for knowledge retrieval. Inspired by \citet{wei2022chain}'s taxonomy of reasoning tasks, we developed the Knowledge-Free Reasoning Dataset (KFRD), which consists of three fundamental reasoning tasks: arithmetic reasoning, symbolic reasoning, and logic reasoning. This dataset includes basic reasoning operations for these three fundamental tasks to evaluate a wide range of knowledge-free reasoning cross-lingual transfer performance.

%Due to the issues with existing reasoning datasets, such as the need for knowledge retrieval, we developed the KFRD.  KFRD consists of three fundamental reasoning tasks: arithmetic reasoning (e.g., addition, subtraction, and other mathematical operations), symbolic reasoning(e.g., deletion, reordering, and other symbolic operations), and logic reasoning(e.g., Implication Elimination and other basic logical rules) . It is designed to evaluate a broad spectrum of knowledge-free reasoning and cross-lingual transfer performance.

Inspired by \citet{wei2022chain}'s taxonomy of reasoning tasks, we developed the KFRD, which consists of three fundamental reasoning tasks:  arithmetic reasoning (e.g., addition, subtraction, and other mathematical operations), symbolic reasoning(e.g., deletion, reordering, and other symbolic operations), and logic reasoning(e.g., Implication Elimination and other basic logical rules) . It is designed to evaluate a broad spectrum of knowledge-free reasoning and cross-lingual transfer performance.

We utilized GPT-4~\cite{achiam2023gpt} to generate multilingual parallel templates and fictitious entities, followed by manual verification. We then used code to generate the dataset based on these templates, entities, and predefined rules. This approach ensures that the tasks can be completed without requiring additional knowledge and guarantees the correctness of the QA pairs. The templates are multiple-choice questions, each composed of one input part, one transformation rule, and one options part. The examples and template are provided in Table~\ref{tab:exampleKFRD} and Figure~\ref{fig:template}.
%%% 这一块需要描述模板的细节有输入输出之类的吗？不说是不是也可以看懂

% \subsubsection{Adapted datasets for fine-grained reasoning}
We also use the ASDiv~\cite{miao2021diverse}, Coin Flip~\cite{wei2022chain}, and ProofWriter~\cite{tafjord2020proofwriter} dataset as a representation of arithmetic, symbolic, and logical reasoning to further validate our conclusions. % We carefully choose these datasets because they require minimal knowledge retrieval and have appropriate difficulty, which meets the requirement of evaluating knowledge-free reasoning.
\input{tab/exampleKFRD}

\subsection{Evaluation metric}
% In order to measure the model's cross-lingual transferability while mitigating the difficulty difference in of the datasets, we calculate the Cross-lingual Transfer Ratio (XLTR) following the prior work \cite{gao2024multilingual}.
%%% 感觉we fine-tune the model on source language (trained language) and then evaluate the model on target language (transferred language) 没有必要，是不是在很多地方都介绍过了
% In order to measure the model's cross-lingual transferability, we fine-tune the model on source language (trained language) and then evaluate the model on target language (transferred language).
In order to assess the model's cross-lingual transferability, we select the Cross-lingual Transfer Ratio (XLTR) as the evaluation metric, following \citet{gao2024multilingual}.
The formula is as follows:
\[
\text{XLTR}(s,t)=(\frac {|C_s \cap C_t|} {|C_s|} - A_r) / (1 - A_r)
\]
where \( s \) and \( t \) denote the source and target languages in the transfer. \( C_x \) represents the set of correct answers in language \( x \), and \( A_r \) is the accuracy of random choices for the given task.

% If after training on a task in one specific language (namely, English), the model shows an XLTR score close to 100\% in other untrained languages, we say it achieves fully cross-lingual transfer in this task.
If the model shows an XLTR score close to 100\% in a language direction, we say it achieves fully cross-lingual transfer in this direction.

We also evaluate the accuracy of models before fine-tuning on all datasets and find poor performance, suggesting that most of the model's ability on transferred languages stem from cross-lingual transfer.

\section{Experiment Settings}
\subsection{Language and model choice}
% \vspace{-0.1cm}
\paragraph{Language choice}
To capture linguistic diversity, we selected ten languages based on their extensive use and representation of diverse linguistic families, following \citet{gao2024multilingual}. The languages selected are English (en), German (de), French (fr), Italian (it), Russian (ru), Polish (pl), Arabic (ar), Hebrew (he), Chinese (zh), and Japanese (ja).
Unless specified, we fine-tune the model in English and evaluate it on other languages.
Further details are provided in Appendix \ref{sec:appendix-lang}.

% \vspace{-0.1cm}
\paragraph{Model choice}
% %%% 是不是直接说对于部分实验，我们使用LLaMA-2-7B-Chat as a representative model for analysis，因为好像strategyqa哪一步也是llama2 不过certain analyses是不是有点泛？

We selected several LLMs, including LLaMA-2-7B-Chat~\citep{touvron2023llama}, BLOOMZ-MT-7B~\citep{muennighoff2023crosslingual}, Mistral-7B-Instruct-v0.1~\citep{jiang2023mistral}, and Qwen-1.5-7B-Chat~\citep{qwen}, for our experiments. To optimize resource use and demonstrate results clearly, we used LLaMA-2-7B-Chat~\citep{touvron2023llama} as a representative model for some analyses.

% For certain analyses, we used LLaMA-2-7B-Chat~\citep{touvron2023llama} as a representative model.
% We selected several LLMs including LLaMA-2-7B-Chat~\citep{touvron2023llama}, BLOOMZ-MT-7B~\citep{muennighoff2023crosslingual}, Mistral-7B-Instruct-v0.1~\citep{jiang2023mistral}, and Qwen-1.5-7B-Chat~\citep{qwen} to conduct transferability experiments on existing and synthetic datasets. 

% For experiments focusing on the impact of the training language (Section~\ref{sec:train-language-proficiency}) and interpretability (Section~\ref{sec:interpret}), we used LLaMA-2-7B-Chat as a representative model for analysis.

% For the experiments evaluating impact of the target language proficiency in specific directions (Section~\ref{sec:target-lang-exp}), we selected derived models of LLaMA-2-7B and Mistral-7B to conduct experiments on Arabic and Hebrew respectively, both of which are low-resource languages.

\subsection{Fine-tuning and decoding settings}
We perform LoRA fine-tuning \citep{hu2021lora} on all model blocks in all experiments due to the limited computational resources. More details about fine-tuning can be found in Appendix~\ref{sec:app-exp-details}.

For decoding, we use constrained decoding in all experiments to ensure the model generates only the desired options (e.g., Yes/No for StrategyQA, A/B/C/D for KFRD).
% LoRA instead of fully fine-tuning is used for two reasons: First, it can lower the requirement for computational resource; second, the LoRA fine-tuning shows equivalent or even better performance than full fine-tuning in our preliminary study.
% \subsection{Decoding settings}
% In all experiments, we perform constrained decoding to prevent the model from generating tokens other than the desired choices (e.g., Yes/No for StrategyQA, A/B/C/D for KFRD).

\section{Results}

\begin{figure}[h!]
% \vspace{-0.2cm}
    \centering
    \includegraphics[width=0.3\textwidth]{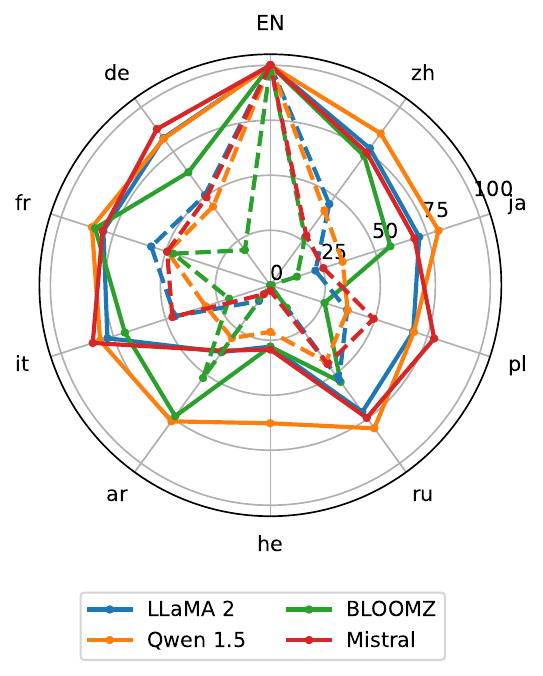}
    \caption{XLTR of different models on StrategyQA. Solid lines: WF-all results; Dashed lines: NF results. The label of training language (en) is capitalized.}
    \label{fig:sqa-xltr}
% \vspace{-0.2cm}
\end{figure}

\begin{figure}[h!]
% \vspace{-0.2cm}
    \centering
    \includegraphics[width=0.3\textwidth]{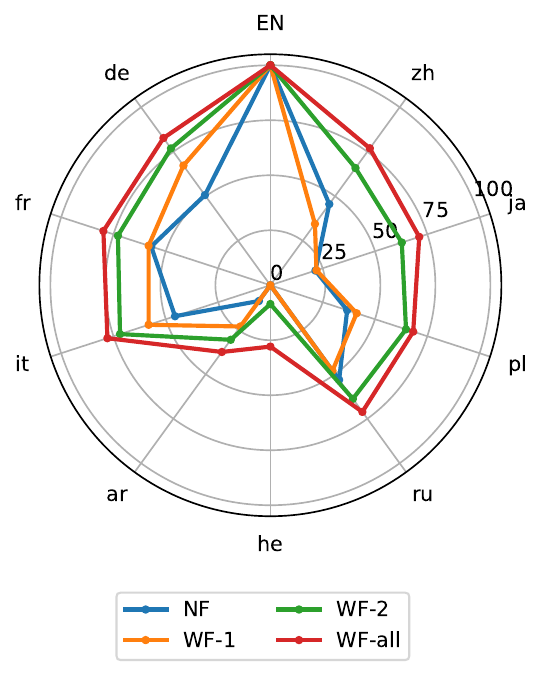}
    \caption{XLTR of LLaMA-2-7B-Chat on StrategyQA under different settings.
    % NF stands for the No Facts setting, while WF-1 and WF-2 indicate that one or two facts are given for each question, respectively. WF-all signifies that all facts are provided.
    }
    \label{fig:sqa-llama2-xltr}
% \vspace{-0.2cm}
\end{figure}

\begin{figure*}[th]
% \vspace{-0.2cm}
    \centering
    \includegraphics[width=0.87\textwidth]{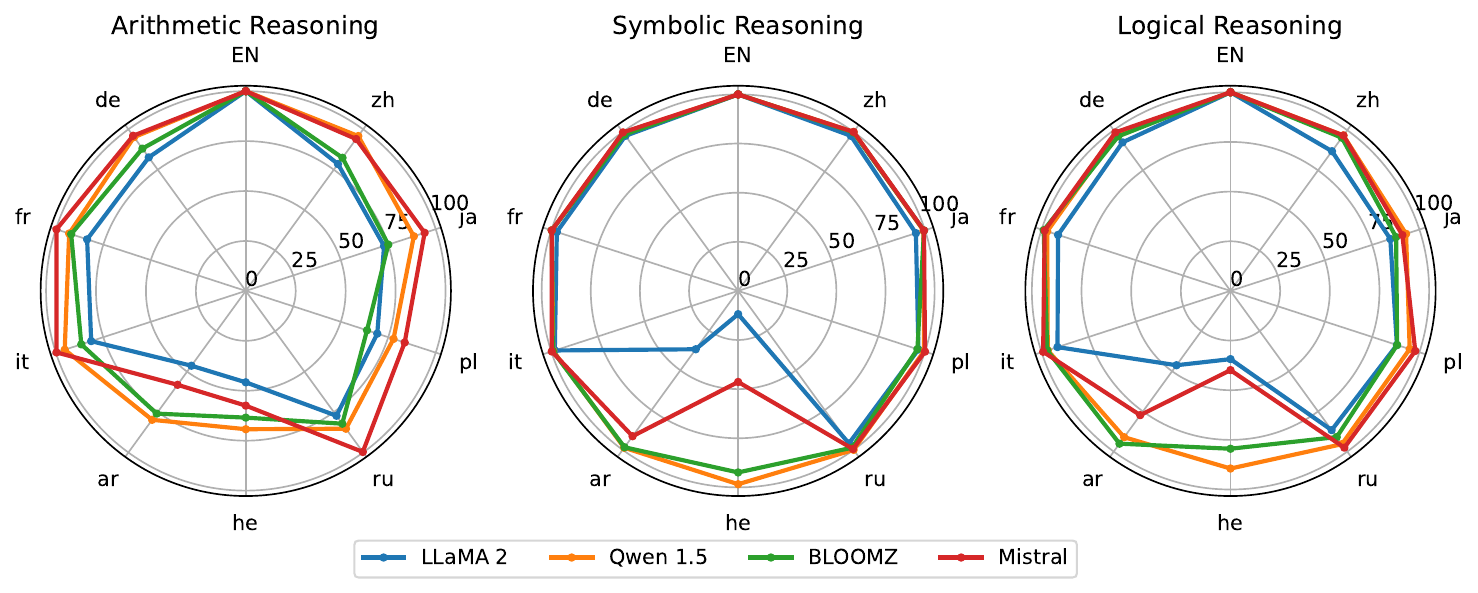}
    \caption{XLTR on the different parts of KFRD}
    \label{fig:main-xltr}
% \vspace{-0.2cm}
\end{figure*}

\begin{figure*}[th]
% \vspace{-0.2cm}
    \centering
    \includegraphics[width=0.87\textwidth]{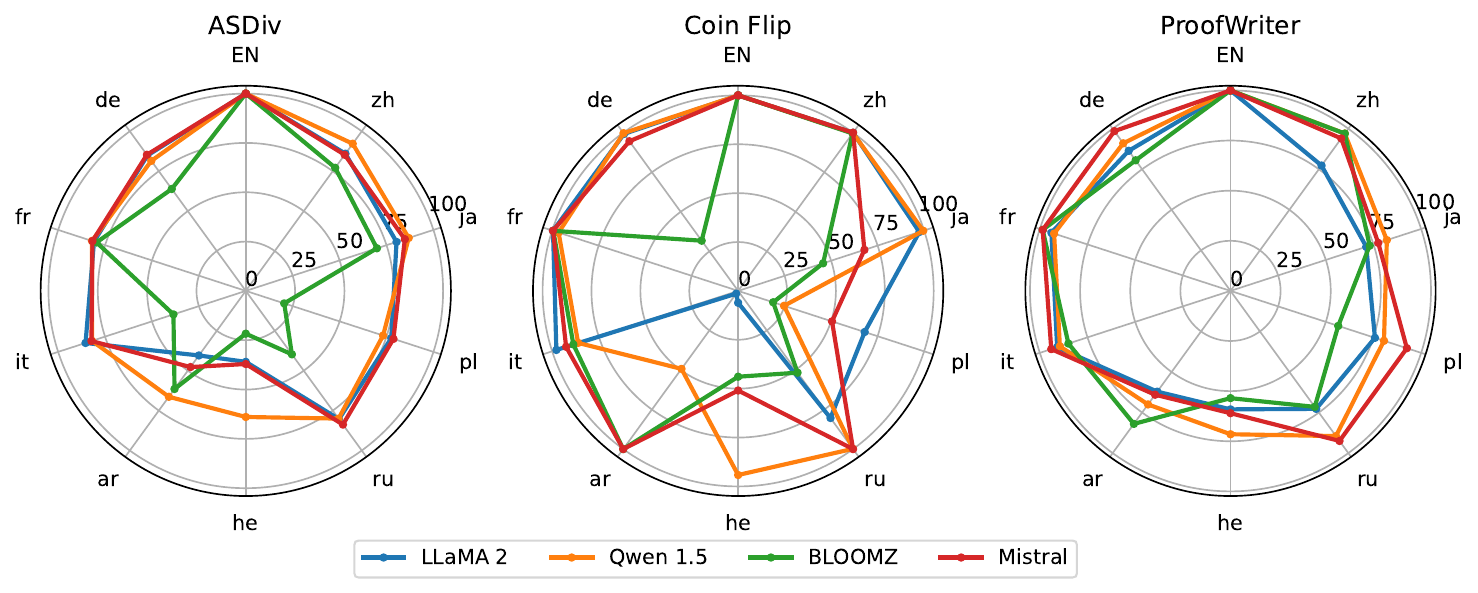}
    \caption{XLTR on the existing pseudo knowledge-free reasoning datasets}
    \label{fig:main-exists-xltr}
% \vspace{-0.2cm}
\end{figure*}

\begin{figure*}[th]
% \vspace{-0.2cm}
\centering
\includegraphics[width=0.87\textwidth]{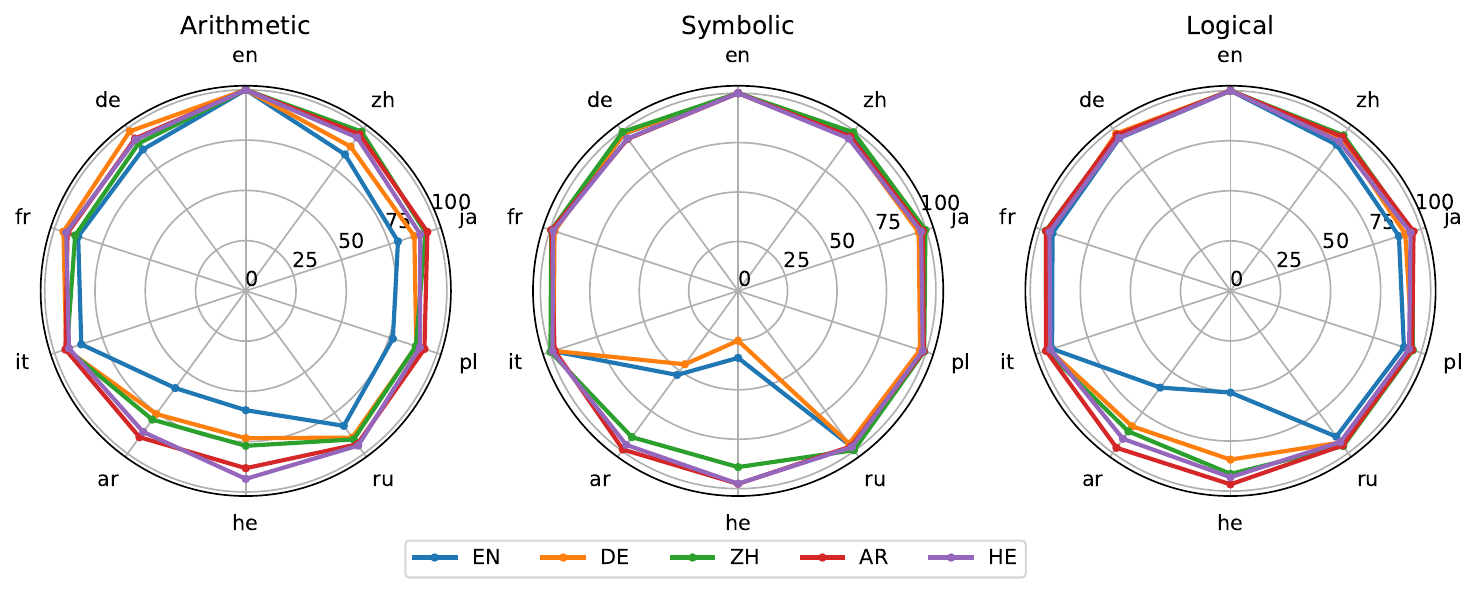}
\caption{XLTR of LLaMA-2-7B-Chat on three parts of KFRD. The different lines indicate different trained languages.}
\label{fig:langs-xltr}
% \vspace{-0.2cm}
\end{figure*}

\subsection{Impact of knowledge retrieval demand on cross-lingual transfer}
We analyze the impact of the amount of knowledge retrieved on cross-lingual transfer in different settings of the reasoning dataset. The results of StrategyQA for the cross-lingual transfer ratio are presented in Figure~\ref{fig:sqa-xltr}, while the accuracy results are detailed in Figure~\ref{fig:sqa-acc}.

% \vspace{-0.1cm}
\paragraph{Knowledge retrieval requirement harms cross-lingual transfer}
The experimental results indicate that, for all languages, the cross-lingual transfer ratio of all models are significantly higher when the necessary knowledge for reasoning is provided compared to when it is not.
This suggests that the requirement for knowledge retrieval significantly hinders the model's cross-lingual transferability when solving reasoning tasks.

%%% 从less修改成了more
% \vspace{-0.1cm}
\paragraph{More knowledge retrieval lowers cross-lingual transfer}
We further conduct detailed evaluations using the LLaMA-2-7B-Chat model to observe the changes in cross-lingual transfer ratios as the amount of knowledge provided varies. As shown in Figure~\ref{fig:sqa-llama2-xltr}, the experimental results demonstrate that the transfer ratio decreases as the demand for knowledge retrieval increases.
This further validates the conclusion that the retrieval of more knowledge significantly impacts cross-lingual transferability.

The results on the QASC dataset were consistent with those mentioned above. Detailed results can be found in Figure~\ref{fig:qasc-xltr} and \ref{fig:qasc-llama2-xltr}.% in Appendix~\ref{sec:app-additional-results}.

\subsection{The cross-lingual transfer of knowledge-free reasoning}
We assess the cross-lingual transferability of the model's knowledge-free reasoning capabilities by evaluating the performance on KFRD and three corresponding existing reasoning datasets.
The resulting cross-lingual transfer ratios are shown in Figures~\ref{fig:main-xltr} and \ref{fig:main-exists-xltr}, while the accuracy results are presented in Figures~\ref{fig:main-acc} and \ref{fig:main-acc}.

The results demonstrate that the KFRD exhibits extremely high cross-lingual transfer performance for most language pairs. For 7 out of the 9 languages, it can be observed that the cross-lingual transfer ratio in knowledge-free reasoning tasks often exceeds 90\%, with some instances approaching 100\%, thus achieving near-full cross-lingual transfer.
Moreover, results from three existing datasets further validate this finding, showing that all models achieve satisfactory transfer ratios across high-resource languages.

For some low-resource languages, such as Hebrew and Arabic in LLaMA-2, German and Hebrew in BLOOMZ~\footnote{For the language distribution, please see Appendix~\ref{sec:app-lang-dist}.}, the cross-lingual transferability is significantly poorer. We hypothesize that this may be due to the model's weaker language proficiency in these languages, which negatively impacts its transferability. Further analysis of this issue is provided in the following section.

% ### TODO：这里的描述得根据结果图修改一下
% 三个现存数据集也展示出了较高的传递率，也可以有力的说明knowledge-free reasoning 可以achieving near fully cross-lingual transfer. 其中coin Flip也是通过代码生成的数据，并可以直接翻译生成模板，所以在高资源语言对上和KFRD性能基本一致。ASDiv由于包含了少量知识，所以准确率会低于KFRD，但仍较高。ProofWriter由于使用了现存的实体，并结合翻译问题，导致检索的知识在一些情况下可能会干扰准确性，导致在一些语言对上略低于KFRD。当排除这些情况时，也可以实现接近fully cross-lingual transfer的性能。我们将具体包含知识的样例，额外的实验结果和详细的描述放在了附录A。   这些特别低的都是在训练语料中没有的，感觉需要加 The three existing datasets also demonstrate high transfer ratios in high-resource languages, strongly supporting the conclusion that knowledge-free reasoning can achieve near-full cross-lingual transfer.

It is noticeable that there are still minor differences in XLTR between KFRD and the existing datasets in the arithmetic reasoning and logical reasoning tasks. However, these differences do not affect the overall conclusion.

We manually check the data samples and find that there are some interfering cases that can affect cross-lingual transfer, such as minor knowledge retrieves, translation issues, and counterfactual information, as detailed discussed in the Appendix~\ref{sec:app-why-create}.

% The three existing datasets also demonstrate high transfer ratios in high-resource languages. For LLaMA and BLOOMZ, the languages with lower transfer ratios have a correspondingly low representation in the training data\citep{touvron2023llama,muennighoff2023crosslingual}, while Qwen and Mistral do not provide details on the distribution of training data. This strongly supports the conclusion that knowledge-free reasoning can achieve near-full cross-lingual transfer. Among them, Coin Flip, which is generated through code and directly translated to create templates, shows similar performance to KFRD across high-resource language pairs. ProofWriter, utilizes existing entities and introduces translation challenges, leading to cases where retrieved knowledge might interfere with accuracy. As a result, its performance is slightly lower than KFRD in some language pairs. However, when these interfering cases are excluded, ProofWriter can also achieve near-full cross-lingual transfer performance. ASDiv, which involves minimal knowledge retrieval, exhibits slightly lower accuracy than KFRD but still performs well. We have provided specific examples of knowledge involvement, additional experimental results, and detailed descriptions in Appendix A.

We also evaluate the LLaMA-2-7B-Chat model on MMLU before and after finetuning, in order to address the concerns of over-fitting on the fine-tuned datasets and forgetting the world knowledge, which is detailed in Appendix~\ref{sec:app-eval-mmlu}.

\subsection{Impact of language proficiency on cross-lingual transfer}
\label{language-proficiency}
%Previous results show that the cross-lingual transfer of knowledge-free reasoning could be related to the proficiency in the transferred language. To further investigate this, 
% In this section, we explore the impact of language proficiency of the training and target language with additional experiments.

\subsubsection{Training language proficiency}
\label{sec:train-language-proficiency}
To evaluate the impact of training language proficiency, based on the data distribution of LLaMA-2 (see Appendix~\ref{sec:app-lang-dist}) and previous experiments, we selected German and Chinese as representatives of high-resource languages, and Arabic and Hebrew as representatives of low-resource languages for training.
Then, we train models on the KFRD in these languages and evaluated their performance across the 10 languages.
The resulting cross-lingual transfer ratios are presented in Figure~\ref{fig:langs-xltr}, while the accuracy results are shown in Figure~\ref{fig:langs-acc}.% in Appendix~\ref{sec:app-additional-results}.

The results show that the models show no significant differences in transfer ratio when trained with high-resource or low-resource languages, indicating that the proficiency and resource of the training language has no significant effect on the cross-lingual transfer of knowledge-free reasoning.

\subsubsection{Target language proficiency} %  in specific directions
\label{sec:target-lang-exp}
In previous experiments, we observe the transferability from English to Arabic and Hebrew was significantly weaker in LLaMA-2 and Mistral.
% In previous experiments, we observe strong cross-linguistic transfer performance between most languages. However, the transferability from English to Arabic and Hebrew was significantly weaker in LLaMA-2 and Mistral.
We hypothesize that this is related to the model's weaker language proficiency in these two target languages.

In this section, we select models from Hugging Face that have undergone Continual Pre-Training (CPT), Supervised Fine-Tuning (SFT), and a combination of both (CPT + SFT) on the LLaMA-2 or Mistral platforms. These adapted models have better proficiency in the respective languages.
The selected models are listed in Table~\ref{tab:training_models}. %in Appendix~\ref{sec:app-target-lang-models}.

\begin{figure}[th]
% \vspace{-0.2cm}
\centering
\includegraphics[width=0.45\textwidth]{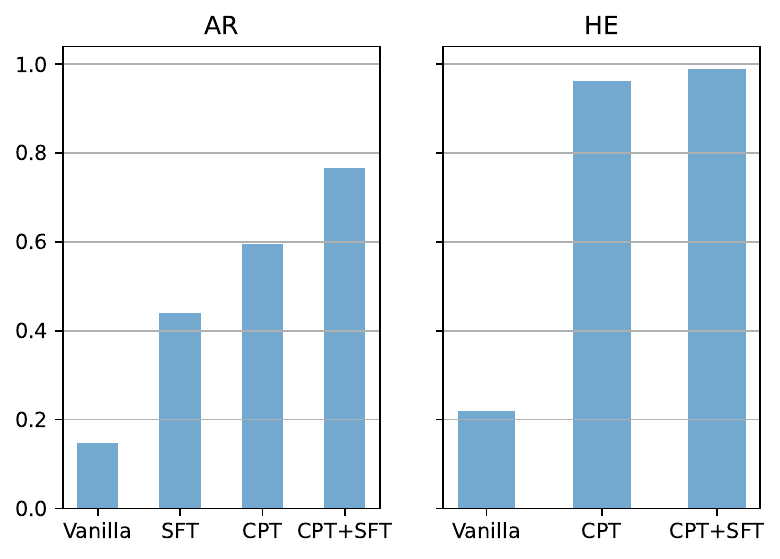}
\caption{Averaged XLTR from English to Arabic/Hebrew across three parts of our KFRD dataset for models in different stages trained in Arabic/Hebrew}
\label{fig:cptft-xltr}
% \vspace{-0.2cm}
\end{figure}

% We train a Vanilla model and the above fine-tuned models on the KFRD in English and evaluate their transferability from English to the respective target languages.
The transfer ratio results of the vanilla and the above fine-tune models are shown in Figure~\ref{fig:cptft-xltr}, and the accuracy results are provided in Figure~\ref{fig:cptft-acc}.%  in Appendix~\ref{sec:app-additional-results}.
The results indicate that the vanilla model exhibits very low transfer rates for low-resource languages. However, after applying CPT, SFT, or CPT+SFT, the transfer ratio increases significantly. Notably, for Hebrew, the transfer ratio reach over 95\%, achieving fully cross-lingual transfer. This suggests that proficiency in Arabic and Hebrew limits the cross-linguistic transfer of the knowledge-free reasoning component, while improving proficiency in the target language can alleviate this limitation.
% This demonstrates that the cross-linguistic transfer of reasoning component in specific directions is influenced by target language proficiency.

%This suggests that the language proficiency in Arabic and Hebrew limits the transfer of knowledge-free reasoning abilities from English to these languages. Improving language proficiency can alleviate this limitation.

%This suggests that the poor cross-lingual transferability for Arabic and Hebrew as target languages is due to insufficient language proficiency. Furthermore, this demonstrates that the target language proficiency significantly impacts cross-lingual transfer performance.

\section{Interpretability Analysis}
\label{sec:interpret}
\subsection{Overview}
Built on previous research~\citep{hu2024limited,gao2024multilingual} and our experiments, we observed that the cross-lingual transferability of knowledge retrieval ability is significantly weaker than that of knowledge-free reasoning. To better understand the reasons behind this difference, we conducted a detailed analysis on model internals using two methods: Cosine Similarity of Hidden States and Neuron Activation.
Both of the methods have been widely used to measure text similarity~\citep{reimers2019sentence,malkiel2022interpreting,wang2024llmcheckup} and analyze models~\citep{dalvi2019neurox,sajjad2022neuron,rai2024investigation}.
% Cosine similarity has been widely used to measure text similarity in various applications \citep{reimers2019sentence,malkiel2022interpreting,wang2024llmcheckup}, while analyzing models through Neuron Activation has also been broadly applied in multiple studies \citep{dalvi2019neurox,sajjad2022neuron,rai2024investigation}.
 
\subsection{Interpretability measurements}
This section introduces the measurements used for interpretability analysis. Further details for these metrics are in Appendix \ref{sec:KFRDadjust}.
% We hypothesize that this is because knowledge-free reasoning is embedded in some language-shared mechanisms, while knowledge is stored separately in different languages.
% To validate this hypothesis, we calculated the similarity of hidden states and the activation of neurons in the Feed-Forward Network. Below is an explanation of these metrics, with more detailed analyses provided in Appendix \ref{sec:cosDetails}.
\subsubsection{Cosine similarity of hidden states (CS)}
We measure the cosine similarity of the hidden representations across multiple languages during the reasoning process of a same question, in order to observe how the semantic space in the tested languages approximate each other. The similarity is calculated by:

% The calculation method is as follows: First, we extract the hidden states of the last token in every question at each model layer. Then, we compute the average cosine similarity for each layer across all language pairs. 
\[
\text{CS}(x) = \frac{\sum_{n=1}^{N} \sum_{\substack{a, b \in \mathcal{L}, a \neq b}} \frac {\mathbf{h}_n^a(x) \cdot \mathbf{h}_n^b(x)} {\left|\mathbf{h}_n^a(x)\right|\cdot\left| \mathbf{h}_n^b(x)\right|}}{|\mathcal{L}|(|\mathcal{L}|-1)N} 
\]
where $x$ is a certain question sample, $N$ is the total number of model layers, $\mathcal L$ denotes the set of all tested languages, and $\mathbf{h}_n^a(x)$ is the output hidden states of the $n$-th layer for sample $x$ in language $a$. After that, the cosine similarity of all tested samples are averaged to report the final score.

\subsubsection{Neuron Activation Overlap (NAO)}
Neuron Activation Overlap measures the extent of shared neuron activations across languages for the same input. 

% To calculate this, we input a question in multiple languages, extract the activation values of the last token of the input, and identify activated neurons based on a threshold for absolute activation values.
% Then NAO is computed as follows for a question sample $x$:
To calculate NAO, we input a question in multiple languages, extract the neuron activation values of the last token of the input, and identify the neurons whose absolute values surpass a set threshold, labeling them as activated.
Then their overlap (NAO) is computed as follows for a question sample $x$:
%Neuron Activation Overlap measures the extent of shared neuron activations across languages for the same input. The steps are:

% \begin{enumerate}
% \item Input a question in multiple languages and extract the activation values of the last token in each language.
% \item Identify activated neurons, considering a neuron activated if its absolute activation value exceeds a threshold.
% \item Calculate the overlap of activated neurons across all tested languages.
% \end{enumerate}

% For a question sample $x$, NAO is as follows:
\[
\text{NAO}(x) = \frac{\left|\mathcal{L}\right|\cdot\left| \bigcap_{l \in \mathcal{L}} S^l(x) \right|}{\sum_{l \in \mathcal{L}} \left| S^l(x) \right|}
\]
where
$\mathcal{L}$ is set of languages, and $S^l(x)$ is the set of activated neurons on sample $x$ in language $l$. After that, the NAO of all tested samples are averaged to report the final score.

\subsection{Knowledge retrieval dataset}
% ###
We selected MKQA~\cite{longpre2021mkqa}, BoolQ~\citep{clark2019boolq}, and AmbigQA~\citep{min2020ambigqa} as representative datasets of knowledge retrieval tasks for the interpretability analysis. Most questions in these datasets can be answered through a single instance of knowledge retrieval. Examples of these datasets are shown in Table~\ref{tab:dataSetExamples}. 
% To further compare the cross-lingual transfer of knowledge retrieval and knowledge-free reasoning, three knowledge-involved  reasoning datasets are used in the interpretability analysis. Most questions in these datasets can be answered through a single instance of knowledge retrieval, with relatively weaker reasoning demands. By using these three datasets, we can more effectively illustrate the differences between knowledge-involved  tasks and knowledge-free reasoning tasks.
% To further compare the cross-lingual transfer of knowledge retrieval and knowledge-free reasoning, we selected MKQA~\cite{longpre2021mkqa}, BoolQ~\citep{clark2019boolq}, and AmbigQA~\citep{min2020ambigqa} as representative knowledge retrieval tasks for the interpretability analysis. Most questions in these datasets can be answered through a single instance of knowledge retrieval, with relatively weaker reasoning demands. By using these three datasets, we can more effectively illustrate the differences between knowledge retrieval tasks and knowledge-free reasoning tasks. Examples of these datasets are shown in Table~\ref{tab:dataSetExamples}.
% 简写open domain，不写下面的描述，直接给case？

% To further compare the cross-lingual transfer of knowledge retrieval and knowledge-free reasoning, three knowledge-involved  reasoning datasets are used in the interpretability analysis.
% 这些数据集都是open-domain，模型在预训练阶段可能见过 These datasets are open domain, and the relevant knowledge is probably trained by the model during pretraining, so these datasets can provide analysis of the model before fine-tuning.

\subsection{Interpretability results}
\subsubsection{Overall computational similarity}
\begin{figure}[th]
% \vspace{-0.2cm}
\centering
\includegraphics[width=0.48\textwidth]{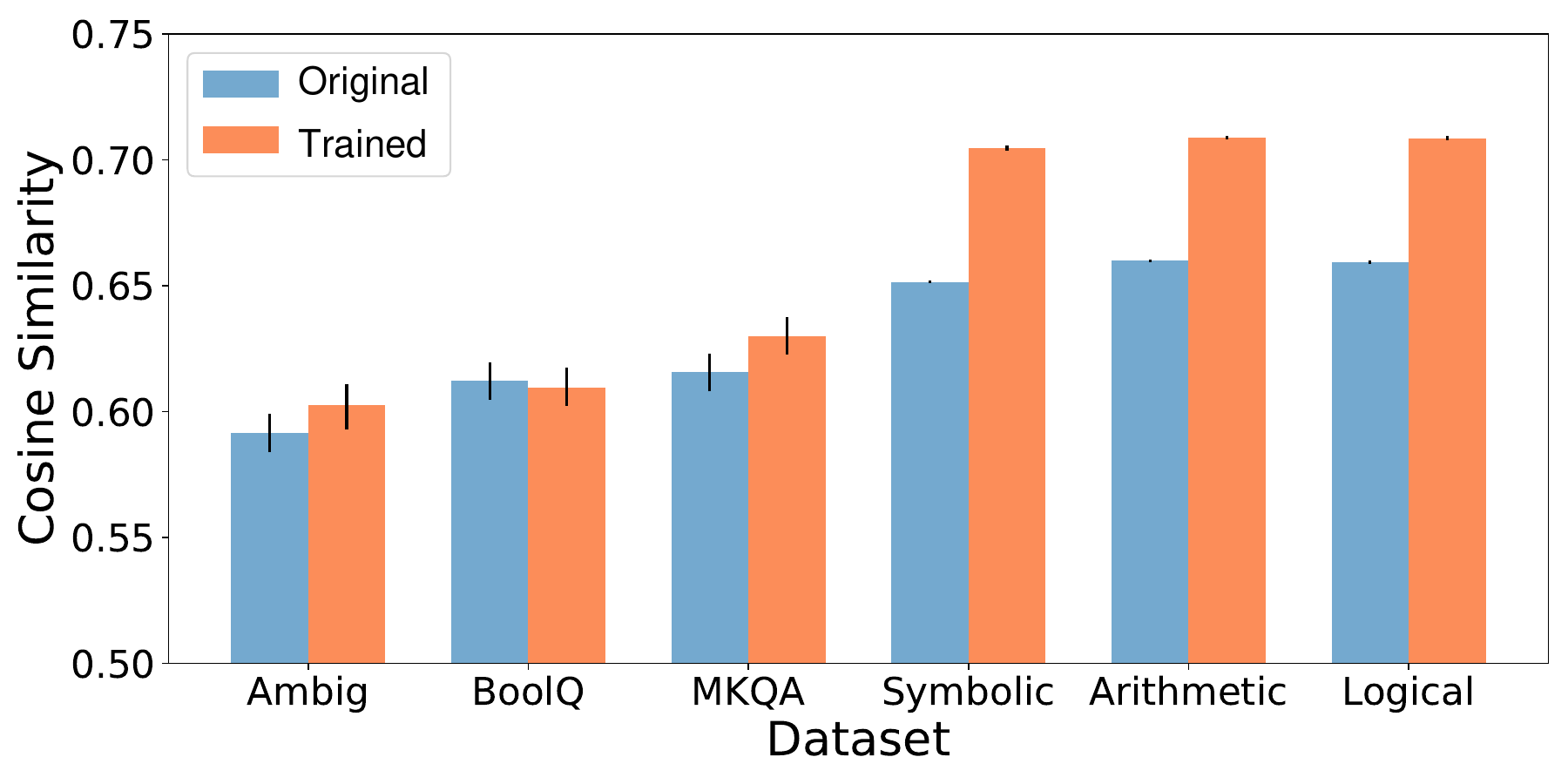}
\caption{CS for different datasets in the LLaMA-2-7B-Chat model. Black lines on each bar indicate the 99\% confidence intervals estimated with bootstrap sampling~\citep{efron1992bootstrap}.} % `Trained' refers to the Fine-tuned model (Original)
\label{fig:cosresult}
% \vspace{-0.2cm}
\end{figure}

\begin{figure}[th]
% \vspace{-0.2cm}
    \centering
    \includegraphics[width=0.48\textwidth]{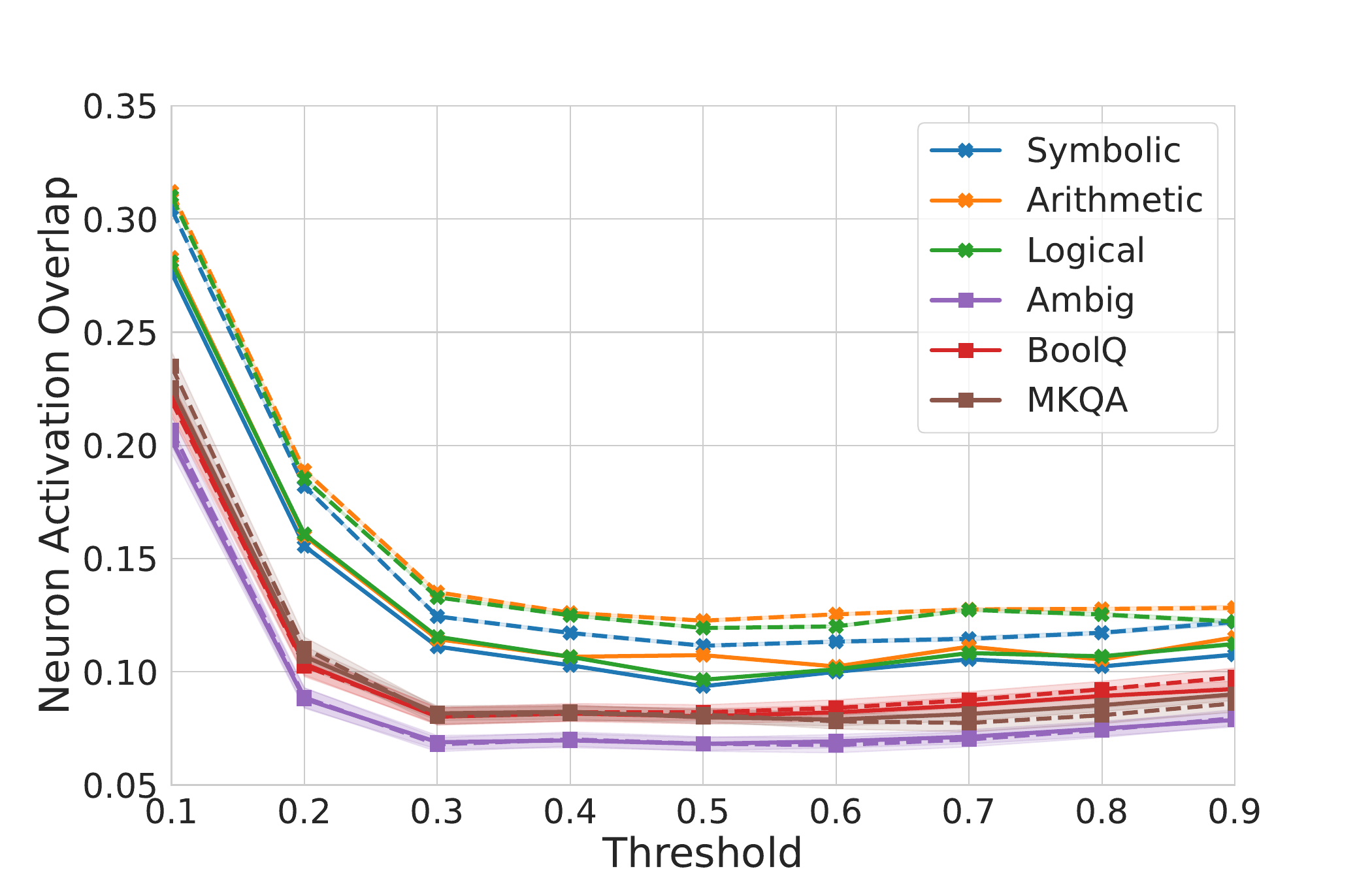}
    \caption{NAO for different dataset in the LLaMA-2-7B-Chat at activation thresholds ranging from 0.1 to 0.9. Shaded areas: 99\% confidence intervals estimated with bootstrap sampling; Solid lines: results of the original model; Dashed lines: results of the LoRA tuned model. The meanings of the shaded areas and dashed lines in Figures~\ref{fig:cosLayerresult} and \ref{fig:actLayerresult} are consistent with those described here.}
    \label{fig:actresult}
% \vspace{-0.2cm}
\end{figure}

% In this section, we evaluate the Feed-Forward Network neural activation overlap and hidden states cosine similarity of the LLama2-7B-Chat model on representative knowledge and reasoning datasets. Additionally, we assess the changes in these two metrics after fine-tuning the model using LoRA on these datasets. The experimental results are presented in Figures \ref{fig:actresult} and \ref{fig:cosresult}, respectively.
In this section, we assess the original and fine-tuned LLaMA-2-7B-Chat model's CS and NAO on knowledge retrieval and knowledge-free reasoning tasks. 
% We also evaluate changes in these metrics after fine-tuning with LoRA to understand the mechanisms supporting cross-lingual transfer of knowledge-free reasoning abilities.
The experimental results are shown in Figures \ref{fig:cosresult} and \ref{fig:actresult}.

% \vspace{-0.1cm}
\paragraph{Internal representation of knowledge-free reasoning task is better aligned than knowledge retrieval}
The results in Figure~\ref{fig:cosresult} indicate that the CS of the model on knowledge-free reasoning tasks is significantly higher than that on knowledge retrieval tasks both before and after fine-tuning.
Additionally, after fine-tuning on knowledge-free reasoning datasets, the CS increases significantly on the corresponding datasets, while fine-tuning on knowledge retrieval datasets shows no significant improvement and may even lead to a decrease. This suggests that adapting to knowledge-free reasoning tasks results in a more aligned hidden space processing across languages.

% Additionally, after training on the knowledge-free reasoning dataset, the model shows a significant increase in CS on the corresponding datasets. In contrast, after training on the knowledge retrieval dataset, the increase in CS is not significant and may even decrease. 
%, whereas adapting to knowledge retrieval tasks leads to relatively divergent hidden space processing.

% \vspace{-0.1cm}
\paragraph{Neuron activation pattern of knowledge-free reasoning task is more similar than knowledge retrieval}
Neuron analysis further elucidates this phenomenon.
The results in Figure~\ref{fig:actresult} show that, across all activation threshold settings, NAO for knowledge-free reasoning tasks is significantly higher than for knowledge retrieval tasks. This indicates that the model tends to use similar neurons for processing knowledge-free reasoning tasks across different languages, resulting in similar neuron activation patterns. Consistent with the hidden states results, after training on the knowledge-free reasoning dataset, NAO increases significantly, whereas there is no significant improvement and even a decline after training on the knowledge retrieval dataset.
This suggests that training on knowledge-free reasoning tasks makes neuron activation characteristics across different languages more similar, leading to the conclusion that the knowledge-free reasoning ability share a similar set of neurons.
% This suggests that training on knowledge-free reasoning tasks makes neuron activation characteristics across different languages more similar.
%, whereas training on knowledge retrieval tasks makes these characteristics relatively divergent.

These results provide a comprehensive analysis of the different cross-lingual transfer effectiveness between knowledge-free reasoning and knowledge retrieval component from a computational similarity perspective. We hypothesize that this difference is because the model stores knowledge for different languages in different neurons, while using similar neuron groups for knowledge-free reasoning.

% %%%以下为原文
%These results provide a comprehensive analysis of the differing cross-lingual transfer effectiveness between knowledge-free reasoning and knowledge retrieval from a computational similarity perspective. We hypothesize that this difference is because the model tends to store knowledge for different languages in different neurons, while using similar neuron groups for knowledge-free reasoning functions.

\subsubsection{Layer-wise computational similarity}
\begin{figure}[th]
% \vspace{-0.2cm}
    \centering
    \includegraphics[width=0.48\textwidth]{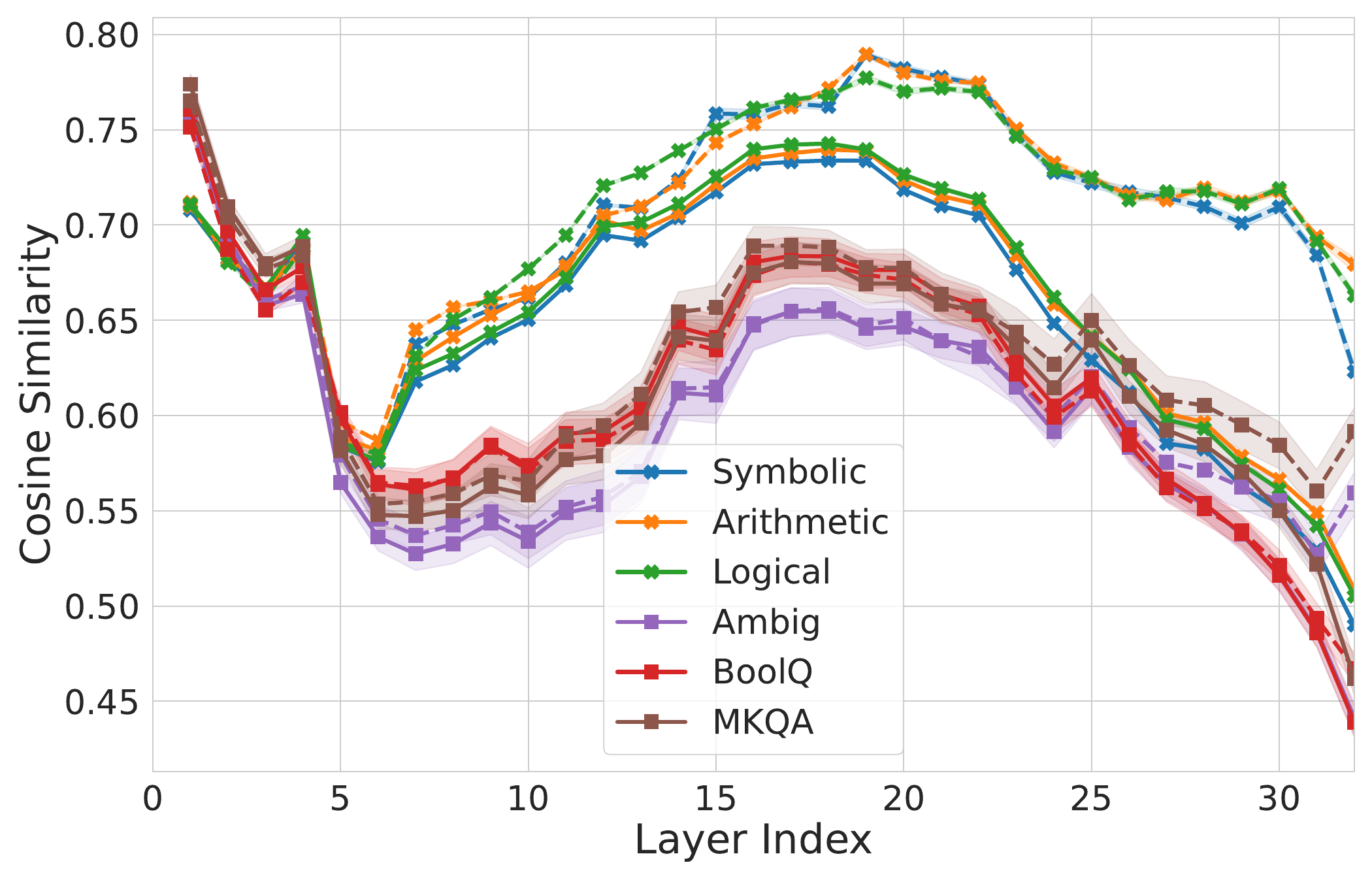}
    \caption{CS for different layers of the LLaMA-2-7B-Chat.}
    \label{fig:cosLayerresult}
% \vspace{-0.2cm}
\end{figure}

\begin{figure}[th]
% \vspace{-0.2cm}
    \centering
    \includegraphics[width=0.48\textwidth]{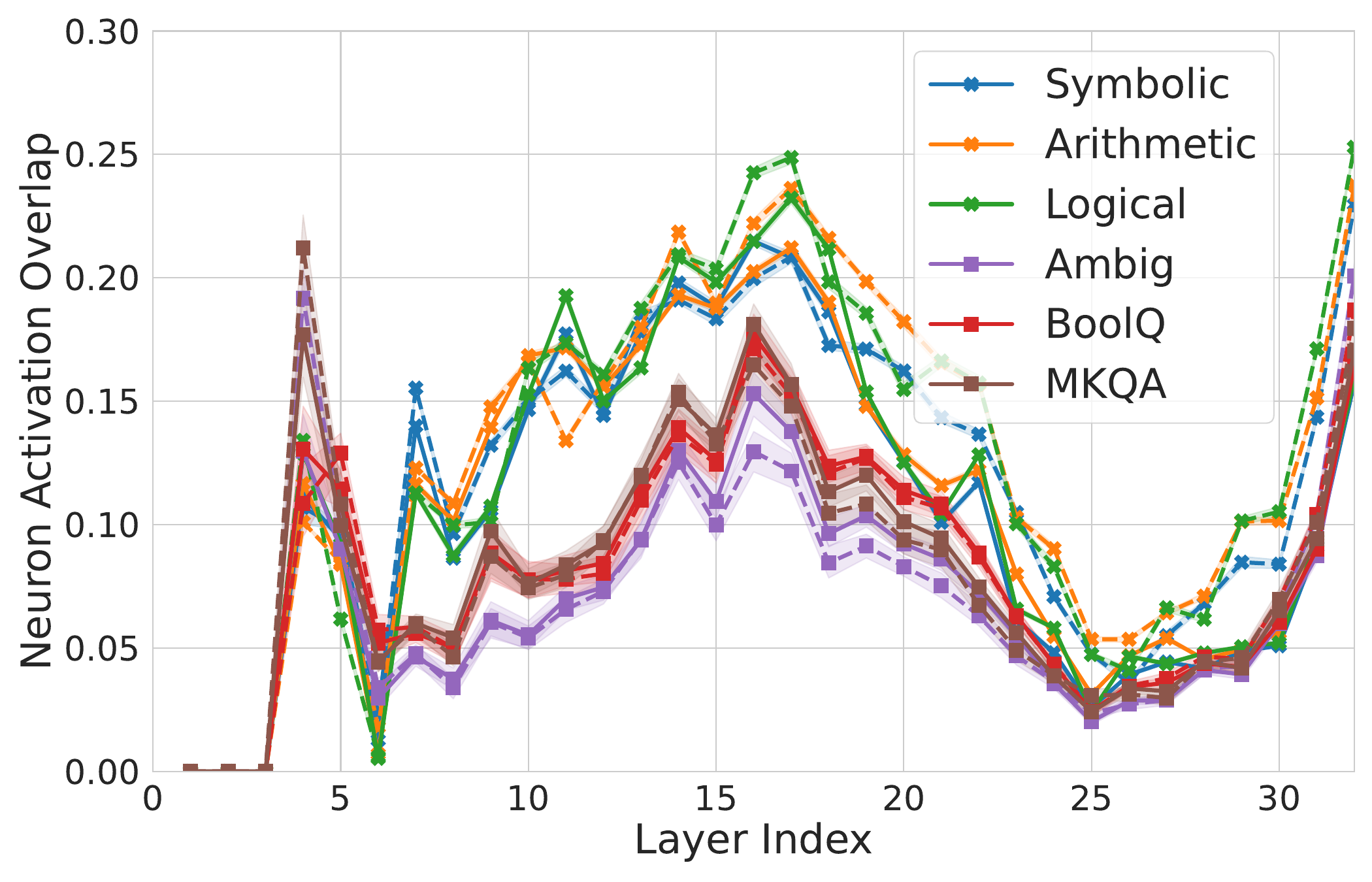}
    \caption{NAO for different layers of the LLaMA-2-7B-Chat at an activation threshold of 0.4.}
    \label{fig:actLayerresult}
% \vspace{-0.2cm}
\end{figure}

To gain finer-grained insights, we performed a layer-wise analysis of CS and NAO. The experimental results are shown in Figures~\ref{fig:cosLayerresult} and \ref{fig:actLayerresult}.
% To analyze the relationship between hidden states and neuron activation characteristics with cross-lingual transfer of reasoning abilities more granularly, we performed a layer-wise analysis of CS and NAO. The experimental results are shown in Figures~\ref{fig:cosLayerresult} and \ref{fig:actLayerresult}.

% In this section, we perform a layer-wise analysis of the neural activation overlap at an activation threshold of 0.4 and the cosine similarity results. The experimental results are presented in Figures \ref{fig:actLayerresult} and \ref{fig:cosLayerresult}.

It is observed that the significantly higher CS and NAO for knowledge-free reasoning tasks, compared to knowledge retrieval tasks, are most pronounced in the middle layers (layers 6-25). 
% It is observed that the phenomena identified in the previous section, specifically the significantly higher CS and NAO for knowledge-free reasoning tasks compared to knowledge retrieval tasks with more pronounced improvements after fine-tuning, are most evident in the middle layers (layers 6-25) of the model and less in the shallow layers (layers 1-6).
Previous work~\citep{zhao2024large,wendler2024llamas} suggested that the middle layers of LLMs are primarily responsible for conceptual reasoning, which is cross-lingual. This hypothesis aligns with our findings and further supports the view that knowledge-free reasoning capabilities can transfer across languages.

%  原文三阶段是understand，solve taske generate
% Additionally, the upper layers (26-32) show similar CS and NAO patterns for both knowledge-free reasoning and retrieval tasks before training. However, after training, there is little change in these metrics for retrieval tasks, while knowledge-free reasoning tasks show significant improvements in both CS and NAO.
Additionally, the upper layers (26-32) show similar CS and NAO patterns for both knowledge-free reasoning and knowledge retrieval tasks before training, but training improvements are only notable in knowledge-free reasoning.
We find that fine-tuning on knowledge-free tasks significantly enhances multilingual accuracy, leading to more consistent outputs. Since the upper layers primarily handle token generation~\citep{zhao2024large,wendler2024llamas}, this consistency improvement results in higher CS and NAO.

\section{Related Work}
\paragraph{Multilingual reasoning evaluation} 
\citet{laskar2023systematic} performed evaluation for multilingual ability of ChatGPT. 
\citet{shi2022language} found LLMs can perform reasoning in multiple languages using CoT, even for those languages with very low resources.
Their analysis mainly evaluated different reasoning tasks, but did not investigate the reasons for performance variations.
% 但他们只是评估了不同的推理任务，并没有对差异的原因更细致的分析

% \vspace{-0.1cm}
\paragraph{Cross-lingual transfer} \citet{gao2024multilingual} evaluated the cross-lingual transferability of models on multiple reasoning datasets, finding significant variations in transfer performance across different datasets.  Furthermore, \citet{hu2024large} found that knowledge transferability remains weak across various settings.
% Building on their conclusions, we extended the analysis by dividing reasoning tasks into knowledge retrieval and knowledge-free reasoning.
Building on their conclusions, we distinguish between the knowledge retrieval and knowledge-free reasoning components and extend the analysis to all reasoning tasks.
Additionally, \citet{ye2023language} assessed the imbalance of knowledge across different languages in LLMs, observing weak cross-lingual transferability of knowledge. \citet{zhu2024question} discovered that training on translated questions can enhance the cross-lingual transferability of reasoning tasks. 

There are also some works focusing on the cross-lingual transfer in the pre-LLM era.
\citet{devlin2018bert} introduced mBERT, advancing cross-lingual transfer by capturing shared linguistic patterns in a unified embedding space, enabling zero-shot transfer without parallel corpora.
Similarly, \citet{conneau2019unsupervised} showed XLM's effectiveness in optimizing multilingual embeddings, improving performance in translation and classification tasks.
\citet{ansell2021composable} proposed composable sparse fine-tuning, selectively fine-tuning sparse parameters across languages to reduce interference and boost performance, especially in low-resource settings, outperforming adapter-based methods in tasks like NER and NLI.

% We include further discussion of related reasoning works in Appendix \ref{sec:addWorks}.

%我们在他们的结论的基础上，进一步讲推理分成两个部分，并使用人造数据集得到了更加清晰的结果。
% Additionally, \citet{zhu2024question} discovered that training on translated questions can enhance the cross-lingual transferability of reasoning tasks. \citet{ye2023language} assessed the imbalance of knowledge across different languages in LLMs, observing weak cross-lingual transferability of knowledge.

% \vspace{-0.1cm}
\paragraph{Analysis of multilingual internal representation}
\citet{zhao2024large} analyzed the way LLMs handle multilingualism and suggested a three-phase working pattern, which includes understanding, task solving and generation.
\citet{wendler2024llamas} also arrived at a similar conclusion.
Expanding on their findings, we further analyzed the differences in how LLMs handle reasoning and knowledge tasks across languages.
%我们在他们理论的基础上进一步分析了推理和知识的差异。

\section{Conclusion and Discussion}
In this study, we analyze the reasons behind the differing cross-lingual transfer abilities of LLMs on various reasoning tasks. We divide reasoning tasks into two components: knowledge retrieval and knowledge-free reasoning. Our experiments demonstrated that the demand for knowledge retrieval significantly hinders the cross-lingual transfer performance, while the knowledge-free reasoning ability can be nearly fully transferred between languages. This discrepancy arises because knowledge-free reasoning relies on shared neural mechanisms across languages, while knowledge storage tends to be more language-specific.

% We divide reasoning tasks into two components: knowledge retrieval and knowledge-free reasoning. We adapt several existing reasoning datasets and construct a knowledge-free reasoning dataset for analysis. 

% 为了让LLMs understand, generate, and perform tasks effectively in multiple languages, thereby reducing the digital language gap and making the benefits of LLMs accessible globally. 我们需要克服cross-lingual knowledge transfer的困难。本研究也可以为LLM的多语言语料配比提供依据，当资源有限时，我们应该给增加low-resource languages的知识语料，适当减少reasoning的语料。

Based on these findings, for knowledge, we recommend prioritizing the inclusion of multilingual data in training corpora in the future. For reasoning, emphasis should be placed on the quality of reasoning data rather than the number of languages. Furthermore, for future multilingual analysis, we recommend investigating knowledge retrieval and knowledge-free reasoning components individually to gain more targeted and detailed insights.

\section*{Limitations}
One key limitation of this paper is the model selection and language coverage. In our exploration of language proficiency and interpretability experiments, we primarily rely on the LLaMA-2 model. Additionally, other parts of our research utilize only a few models, which may oversimplify the descriptions of model performance and behavior. In terms of language coverage, although we included ten languages from different language families, this number is still insufficient compared to the thousands of languages globally. This limitation is partly due to our computational resource constraints. With adequate resources, the proposed methods could be extended to other models and languages to further validate our conclusions.

Another limitation of our study is the depth of the interpretability analysis. We aim to investigate whether different knowledge-free reasoning tasks utilize the same neurons and whether knowledge is stored in different neurons for different languages. However, our support for this hypothesis is primarily based on macro-level numerical analyses, without precisely identifying specific reasoning neurons and knowledge neurons. This limitation restricts our fine-grained understanding of the model's internal mechanisms. Future research should conduct more detailed neuron-level analyses to verify these hypotheses.

\section*{Ethics Statement}
The authors declare no competing interests. All datasets used in this study are sourced from publicly available repositories and do not contain sensitive information, such as personal data. The data generated by GPT-4 have been verified to be non-toxic and are used exclusively for research purposes. The use of LLaMA-2 models, as well as several other large language models, complies with their respective licenses.
\section*{Acknowledgements}
We would like to thank the anonymous reviewers for their insightful comments. Shujian Huang is the corresponding author. This work is supported by National Science Foundation of China (No. 62376116, 62176120), research project of Nanjing University-China Mobile Joint Institute, the Fundamental Research Funds for the Central Universities (No. 2024300507).

\bibliography{custom}

\appendix
\renewcommand{\thetable}{A\arabic{table}} 
\setcounter{table}{0} % 从A1开始编号
% 修改图片编号格式，使其包含附录字母 
\renewcommand{\thefigure}{A\arabic{figure}} 
\setcounter{figure}{0} % 从A1开始编号
\section{Details of Dataset}
\label{sec:app-dataset-detail}
\subsection{Detailed description of Knowledge-Free Reasoning Dataset}
\label{sec:KFRDdetails}
The KFRD is generated using a unified template, consisting entirely of multi-choice questions with four options.
We first create parallel templates for 10 languages using GPT-4 and then fill in different parts of the template with pre-defined rules.
Each question is structured into three parts: input, output, and transformation rules. Specific examples can be seen in Table \ref{tab:exampleKFRD}, and the templates used for these examples are shown in Figure \ref{fig:template}.

\subsubsection{Arithmetic reasoning}
This dataset transforms two input numbers through mathematical operations into one or two output numbers. The mathematical operations include addition, subtraction, multiplication, division, equality, geometric progression, arithmetic progression, and sorting. Each of the three parts are generated by the following rules:

\begin{itemize}
    \item \textbf{Input:} Numbers are randomly generated within the range of 0-999.
    \item \textbf{Transformation rules:} Each rule generates an equal number of samples.
    \item \textbf{Output:} Generated through transformation rules, constrained within the range of 0-999. Other options are randomly generated, ensuring a single correct answer.
\end{itemize}

% The training dataset for Arithmetic Reasoning consists of 8000 samples, and the testing dataset consists of 800 samples.

% Due to the limited range, the training dataset for Arithmetic Reasoning consists of 8000 samples, and the testing dataset consists of 800 samples. Expanding the range would increase difficulty and necessitate a substantial increase in data volume, hence the choice of the 0-999 range.

\subsubsection{Symbolic reasoning}
This dataset transforms 3-5 input words from the corresponding language through symbolic operations to generate the output. Symbolic operations include repetition, addition, deletion, reordering, and their combinations. Considering that single-step symbolic operations are too simple, we chose up to three-step symbolic operations. Each of the three parts are generated by the following rules:

\begin{itemize}
    \item \textbf{Input:} Randomly select 3-5 words from a specific language. We chose 100 simple English words and translated them into other languages using Google Translate.
    \item \textbf{Transformation rules:} The dataset includes equal amounts of single-step, two-step, and three-step symbolic operations. For single-step operations, each rule generates an equal number of samples. For two-step and three-step operations, rule combinations are randomly selected.
    \item \textbf{Output:} Generated through transformation rules. Other options are partially randomly generated and partially based on random replacements from the original input, ensuring consistent length and a unique correct answer.
\end{itemize}

% The training dataset for Symbolic Reasoning consists of 2000 samples, and the testing dataset consists of 500 samples.

\subsubsection{Logical reasoning}
This dataset generates output from a subset of eight input propositions using logical rules. Logical rules include Implication Elimination, Conjunction Introduction, Conjunction Elimination, Disjunction Introduction, Disjunction Elimination, and Proof by Contradiction. The Logical rules are referenced from \citet{saparov2024testing}. Each of the three parts are generated by the following rules:

\begin{itemize}
    \item \textbf{Input:} Eight propositions are generated using proposition templates and randomly selected entities, proposition templates referenced from \citet{saparov2024testing} and entities from \citet{saparov2024testing} and \citet{gao2024multilingual}. Missing languages were supplemented using Google Translate.
    \item \textbf{Transformation rules:} Each logical rule generates an equal number of samples.
    \item \textbf{Output:} Generated through logical rules. Other options are partially based on entities appearing in the propositions and partially randomly generated, ensuring a unique correct answer.
\end{itemize}

% The training dataset for Logical Reasoning consists of 4000 samples, and the testing dataset consists of 500 samples.

% \begin{table*}[h]
% \centering
% \renewcommand{\arraystretch}{1.5} % 调整行高
% \begin{tabular}{l m{4cm} m{3cm} m{5.5cm}} % 调整列宽
% \toprule
% \textbf{Task} & \textbf{Input} & \textbf{Transformation Rule} & \textbf{Output Options} \\ 
% \midrule
% \textbf{Arithmetic} & 11,645 (one or two numbers) & Addition (a mathematical operation) & \makecell[l]{A) 595 \\ B) 536 \\ C) 771 \\ \textbf{D) 656} (one number, sometimes the result is two numbers)}  \\
% \midrule
% \textbf{Symbolic} & education, game, president, night, man (3-5 words in the corresponding language) & Swap the positions of the 5th and 2nd words; Delete the 2nd word (symbolic operations within three layers) & \makecell[l]{\textbf{A) education, president, night, game} \\ B) education, problem, night, game \\ C) hand, president, night, game \\ D) education, house, night, game} \\ 
% \midrule
% \textbf{Logical} & Alex is Aurora Vale. Everything that is Aurora Vale is Omicron Delta. Stella is not Chronos Wasteland. Max is not Dreamweaver's Haven. Suppose Sally is Whispering Meadows, then Sally is Chimerical Citadel. Everything that is Ebonwyrm Abyss is Phoenixfire Ridge. (8 propositions) & Implication Elimination (a logical rule) & \makecell[l]{A) Alex is Seraphim Heights. \\ B) Alex is Tempestwilds. \\ \textbf{C) Alex is Omicron Delta.} \\ D) Polly is Arcadia Reach.} \\ 
% \bottomrule
% \end{tabular}
% \caption{Examples of different tasks in the KFRD dataset}
% \label{table:example}
% \end{table*}

\begin{figure}[ht]
\begin{tcolorbox}[colframe=gray!75!black,colback=gray!10!white,arc=4mm,boxrule=0.5mm]
\small \textbf{Instruction:} The output is the result of applying a specific transformation rule to the input. In this question, you will be given an input value and its corresponding transformation rule. Based on this information, determine the correct output from the options provided: A, B, C, or D. Please give the corresponding answer option directly.\\[1ex]
\textbf{Transformation Rule:} \{Transformation Rule\}\\
\textbf{Input:} \{Input\}\\
Based on the above rule and input, choose the correct output from the following options:\\
% \begin{enumerate}[label=\Alph*.]
A. Output: \{Output1\}\\
B. Output: \{Output2\}\\
C. Output: \{Output3\}\\
D. Output: \{Output4\}\\
% \end{enumerate}
\textbf{Answer:}
\end{tcolorbox}
\caption{Example prompt template for our KFRD dataset}
\label{fig:template}
\end{figure}
\begin{table*}[ht]
    \centering
    \footnotesize
    \begin{tabular}{lll}
        \toprule
        \textbf{Training} & \textbf{Arabic} & \textbf{Hebrew} \\
        \midrule
        Vanilla & LLaMA-2-7B-Chat & Mistral-7B-Instruct-v0.1 \\
        SFT & Llama-2-7b-chat-arabic-lora~\citep{icebear2024llama} & - \\
        CPT & SambaLingo-Arabic-Base~\citep{csaki2024sambalingo} & DictaLM-2.0~\citep{dicta2024dictalm2} \\
        CPT+SFT & SambaLingo-Arabic-Chat & DictaLM-2.0-Instruct \\
        \bottomrule
    \end{tabular}
    \caption{Training models for Arabic and Hebrew}
    \label{tab:training_models}
\end{table*}

\subsection{Detail of existing pseudo knowledge-free reasoning datasets}

Here we provide more details on the datasets used in the experiment.
\begin{itemize}
\item For the ASDiv dataset, we use the subset that contains only arithmetic operations (ASDiv-A\footnote{\url{https://github.com/chaochun/nlu-asdiv-dataset/tree/master/dataset/nfolds/asdiv-a}}) for ease of evaluation. We use folds 0-3 for training and fold 4 for testing.
\item For the ProofWriter dataset, we use the depth-1 subset for evaluation considering the appropriate difficulty.
\end{itemize}

% \subsection{Model Performance on Datasets Before Fine-tuning}
% We evaluated the accuracy of models before fine-tuning on these datasets and find that they have poor performance, so we can assume that most of the model's ability on languages besides trained language comes from cross-lingual transfer.

\subsection{Translation process for English-only datasets}
\label{translationQuality}
For datasets available only in English, we translate them into other languages with Google Translate and verify translation quality with GPT-4. 

Google Translate is highly regarded in the field of commercial translation and is widely used in multilingual research \citep{chen2021mtg,ye2023language,omar2024systematic,song2024missing}. To ensure translation accuracy, we sampled a subset of translation results and employed GPT-4 for verification. Due to budget constraints, we were unable to employ human translators.

For the StrategyQA dataset, we utilized Google Translate and conducted a sample check of 100 items for each language using GPT-4. This process resulted in an overall quality score of 4.47 (on a scale of 1-5), which we consider acceptable for our purposes.

\section{Language Choice}
\label{sec:appendix-lang}
This section provides an overview of the languages utilized in our research, highlighting the primary countries where they are spoken and their respective language families. Refer to Table \ref{tab:language_countries} for detailed information.

\begin{table}[ht]
    \centering
    \footnotesize
    \begin{tabular}{ccc}
\hline
\textbf{ISO} & \textbf{Country Samples} & \textbf{Language Family} \\ \hline
en & US, UK & Germanic \\
de & Germany, Austria & Germanic \\ \hline
fr & France, Canada & Romance \\
it & Italy & Romance \\ \hline
pl & Poland & Slavic \\
ru & Russia, Belarus & Slavic \\ \hline
ar & Egypt, Algeria & Afro-Asiatic \\
he & Israel & Afro-Asiatic \\ \hline
ja & Japan & Japonic \\ \hline
zh & China (Mainland) & Sino-Tibetan \\ \hline
\end{tabular}
    \caption{Correspondence between Languages, Country Samples, and Language Families}
    \label{tab:language_countries}
\end{table}

\section{Implementation Details for Interpretability}
\label{sec:KFRDadjust}
\subsection{Calculation method for activation values}
We use the output of the gate linear layer in the SwiGLU module of the LLaMA model, processed through the SiLU function, as the activation values.

\subsection{Reasons for using the last token for analysis}
In the interpretability analysis, we use the last token of the question to collect the hidden states and neural activation values, because the last input token is used to predict the next token, it gradually incorporates the primary information of the entire sentence, reflecting the overall thought process for the entire problem \citep{meng2022locating,stolfo2023mechanistic,wu2024language}. By focusing on the model’s computational pathway for reasoning rather than calculating the similarity between multilingual sentences, we can better understand how the model processes complex queries. Calculating with an output token, on the other hand, would make it difficult to interpret the reasoning process. Additionally, token counts differ across languages, complicating direct comparisons. Therefore, using the last input token helps in standardizing the analysis across different languages. 

% We make some adjustments to these datasets. For more details, see Appendix 
\subsection{Dataset adjustments}
% To ensure consistency in the final token across different datasets, we added a language-specific "?" to certain datasets. For details on these modifications and metrics, please refer to Appendix \ref{sec:KFRDadjust}.
% To ensure consistency in the final token across datasets, we added a language-specific "?" where needed. This standardization eliminates inconsistencies in input tokens, which could lead to unreliable internal representations, particularly in the lower layers of the model. The "?" also serves as a trigger for the model to initiate the response process.

To ensure consistency in the final token across different datasets, we made slight modifications by adding a language-specific "?" where needed.

Since we are analyzing the internal representation of the last token, in this way, we can eliminate interference caused by the inconsistent input token, which may make the representation unreliable, especially in the bottom layers.
Another reason why we append the token "?" is that it can act as a trigger to let the model start the process of preparing to answer the question, which is what we are analyzing.

For knowledge-free reasoning dataset, we added the phrase "Which option should I choose?" in different languages. For the MKQA and BoolQ datasets, where some questions did not end with a "?", we added a "?". All other datasets already ended with a "?".

\begin{table}
\centering
\footnotesize
\begin{tabular}{l|cc}
\hline
\textbf{Dataset} & \textbf{Samples} & \textbf{Epoch}\\
\hline
StrategyQA & 2061 & 4\\
KFRD Arithmetic & 8000 & 4\\
KFRD Symbolic & 2000 & 1\\
KFRD Logical & 4000 & 1\\
\hline
\end{tabular}
\caption{Training epoch and number of samples of fine-tuned datasets in the transferability experiments}
\label{tab:data-eval-epoch-samples}
\end{table}

\begin{table}
\centering
\footnotesize
\begin{tabular}{l|c}
\hline
\textbf{Dataset} & \textbf{Samples} \\
\hline
StrategyQA & 228 \\
KFRD Arithmetic & 800 \\
KFRD Symbolic & 500 \\
KFRD Logical & 500 \\
\hline
\end{tabular}
\caption{The size of testset used in the transferability experiments}
\label{tab:testdata-eval-size}
\end{table}

\section{Experiments Details}
\label{sec:app-exp-details}
This section outlines the details of our experiments for reproducibility.

\subsection{Infrastructure}
We used the following scientific artifacts in our research:
\begin{itemize}
\item \textit{PyTorch}~(\citealp{Ansel_PyTorch_2_Faster_2024}, BSD license), a framework for building and running deep learning models.
\item \textit{Transformers}~(\citealp{wolf-etal-2020-transformers}, Apache-2.0 license), a library providing a user friendly interface for running and fine-tuning pre-trained models.
\item \textit{DeepSpeed}~(\citealp{rasley2020deepspeed}, Apache-2.0 license), a library optimizing the parallel training of the deep learning models.
\item \textit{LLaMA-Factory}~(\citealp{zheng2024llamafactory}, Apache-2.0 license), a library that provides a unifying way to easily fine-tune large language models with parameter efficient fine-tuning technique like LoRA.
\end{itemize}
% We employ PyTorch~\citep{Ansel_PyTorch_2_Faster_2024} and Hugging Face Transformers~\citep{wolf-etal-2020-transformers} to load and execute the LLMs, and we utilize DeepSpeed~\citep{rasley2020deepspeed} library and the LoRA modules from the LLaMA-Factory~\citep{zheng2024llamafactory} repository to fine-tune the models.

\subsection{Hyperparameters}
In the fine-tuning of all models, we use a learning rate of 2e-4 with a cosine learning rate scheduler. We clip the gradient norm to 1.0, use a total batch size of 64, set the rank of LoRA to 128, and alpha to 16. The LoRA adapters are applied to all the linear layers within Transformer blocks.

The numbers of training epoch and samples used in the transferability experiments are listed in Table~\ref{tab:data-eval-epoch-samples}.
These numbers are tuned to enable LLaMA-2-7B-Chat to achieve 85\% + accuracy on the corresponding tasks.
The size of testsets used in the transferability experiments are shown in Table~\ref{tab:testdata-eval-size}.

In the interpretability experiments, we adjust the number of training epochs or the size of the syntactic datasets to keep the number of total update steps on all datasets around 150, which avoids interference of different update steps on experimental results.
We report the average cosine similarity and neuron activation overlap of 100 samples from each data set.
\input{tab/sqaExample}

\subsection{Computation resources}
All the fine-tuning experiments can be done on 4 NVIDIA Tesla V100 32GB GPUs.
Each fine-tuning can be done in no more than 2 hours.

\subsection{Models used in the target language proficiency experiment}
\label{sec:app-target-lang-models}
The continue pre-training or fine-tuning models of LLaMA-2-7B and Mistral-7B used in the target language proficiency experiment in \ref{sec:target-lang-exp} are listed in Table~\ref{tab:training_models}.

\section{Additional Results of Experiment}
\label{sec:app-additional-results}
\begin{figure*}[ht]
\centering
\begin{minipage}{0.32\linewidth}
\centering
\includegraphics[width=\textwidth]{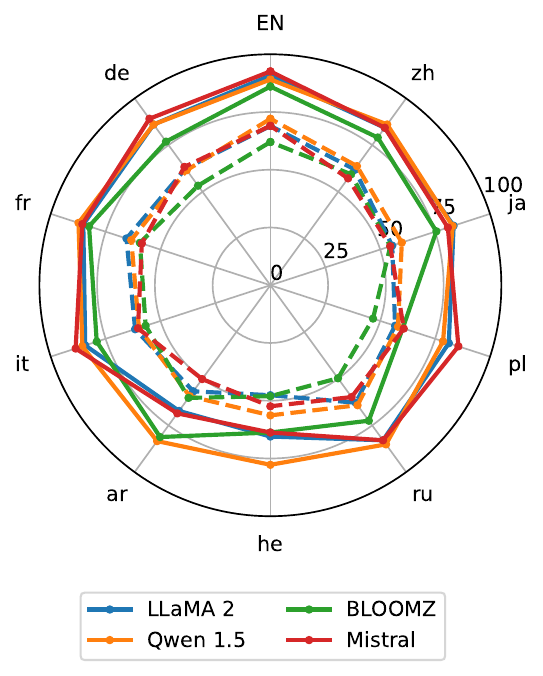}
% \caption{Accuracy of different models on StrategyQA. Solid and dashed line represent the result of With Facts and No Facts setting, respectively.}
\end{minipage}
\begin{minipage}{0.32\linewidth}
\centering
\includegraphics[width=\textwidth]{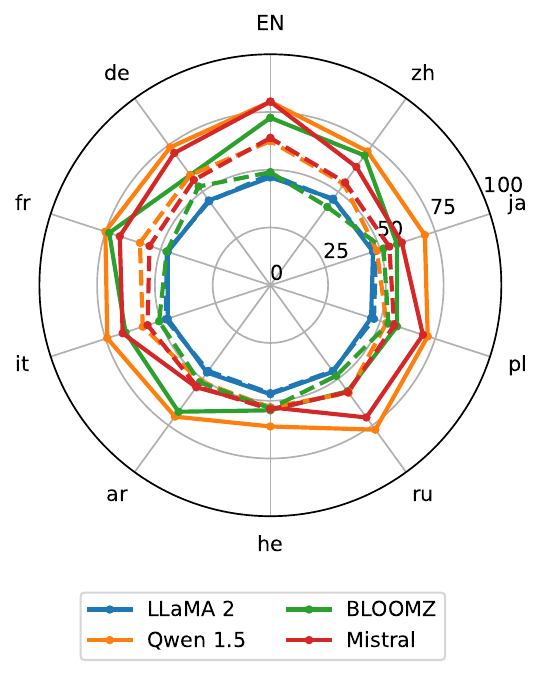}
% \caption{Accuracy of different models on StrategyQA before fine-tuning. Solid and dashed line represent the result of With Facts and No Facts setting, respectively.}
\end{minipage}
\begin{minipage}{0.32\linewidth}
\centering
\includegraphics[width=\textwidth]{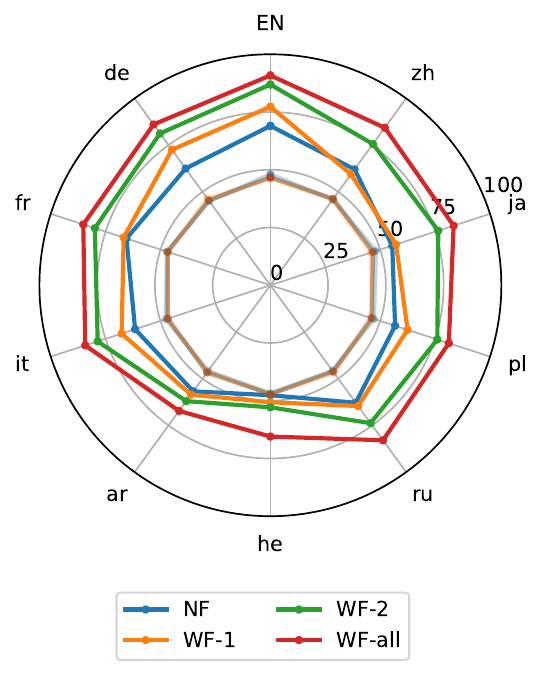}
% \caption{Accuracy of LLaMA 2 7B Chat on StrategyQA under various settings. The translucent line represents the accuracy before finetuning on the specific tasks.}
\end{minipage}
\caption{\textbf{Left}: Accuracy of different models on StrategyQA. Solid and dashed line represent the result of With Facts and No Facts setting, respectively. \textbf{Middle}: Accuracy of different models on StrategyQA before fine-tuning. \textbf{Right}: Accuracy of LLaMA-2-7B-Chat on StrategyQA under various settings. The translucent line represents the accuracy before finetuning on the specific tasks (which are all around 50\%).}
\label{fig:sqa-acc}
\end{figure*}

\begin{figure*}[ht]
\centering
\includegraphics[width=1.0\textwidth]{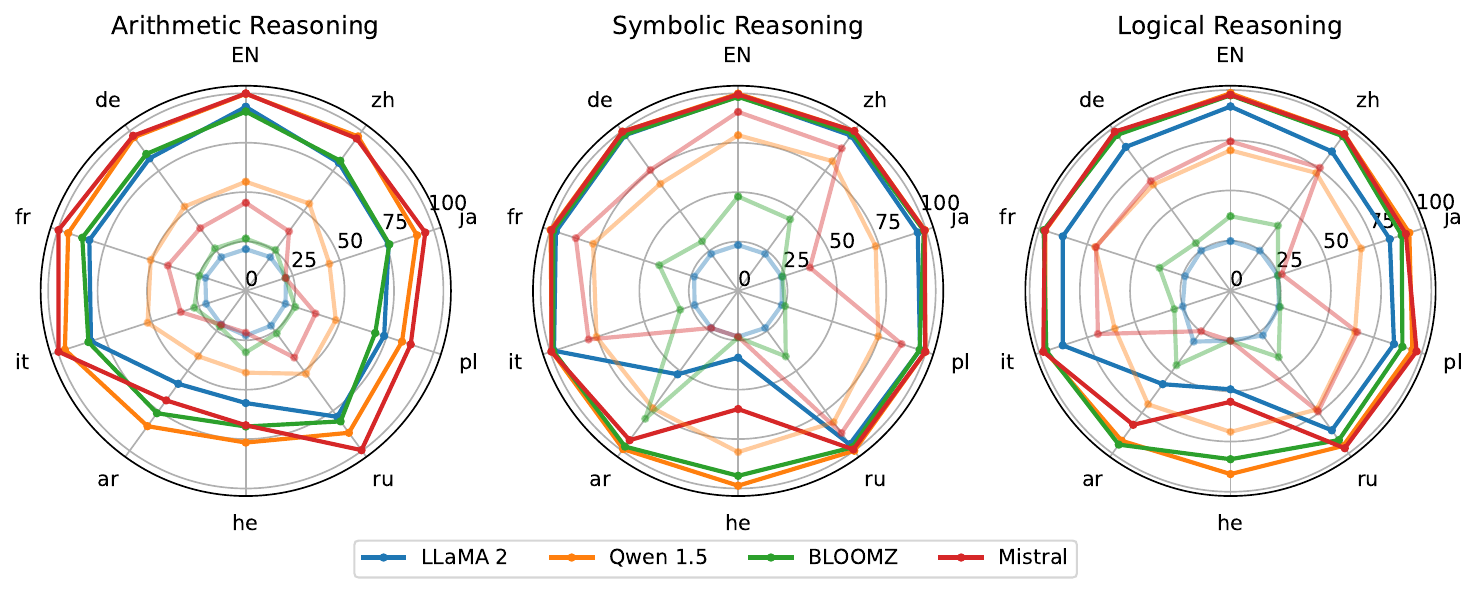}
\caption{Accuracy of various models on different parts of KFRD. The translucent line represents the accuracy before finetuning on the specific tasks.}
\label{fig:main-acc}
\end{figure*}

\begin{figure*}[th]
\centering
\includegraphics[width=1.0\textwidth]{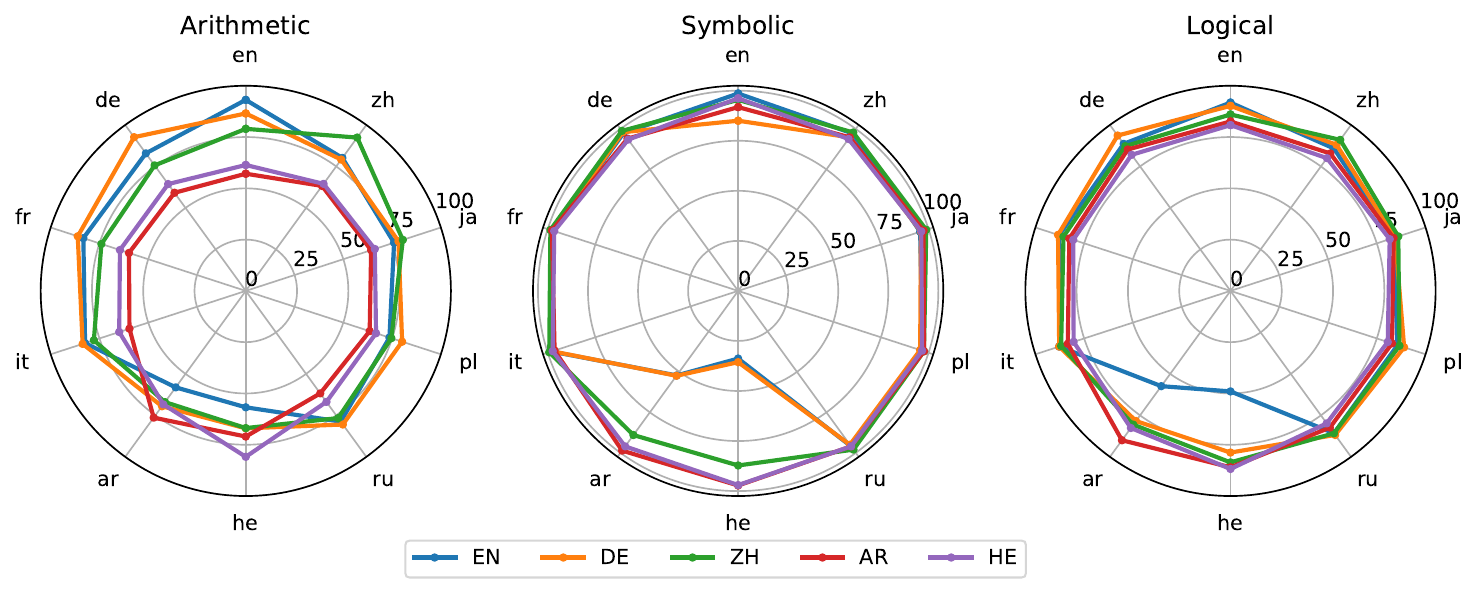}
\caption{Accuracy of LLaMA-2-7B-Chat on three parts of KFRD. The different lines indicate different trained languages.}
\label{fig:langs-acc}
\end{figure*}

\begin{figure}[ht]
\centering
\includegraphics[width=0.45\textwidth]{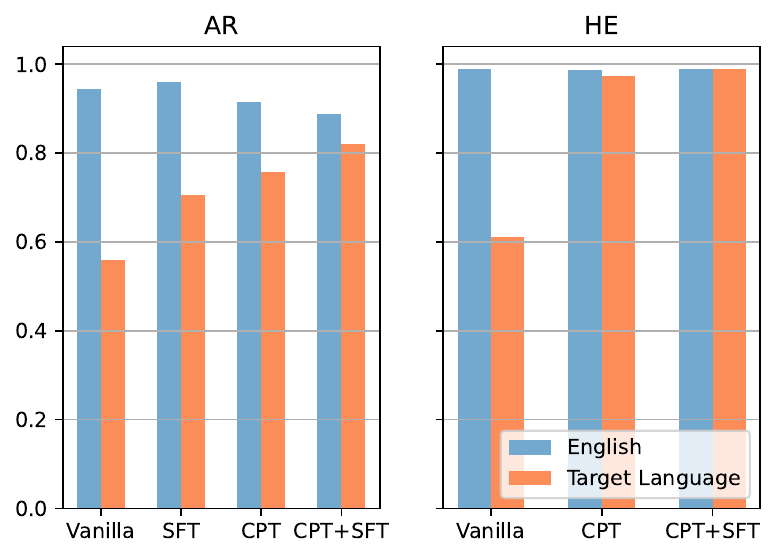}
\caption{Averaged Accuracy on English and Arabic/Hebrew KFRD for models in different stages trained in Arabic/Hebrew}
\label{fig:cptft-acc}
\end{figure}

Here we provide the accuracy of the above experiments in Figure \ref{fig:sqa-acc}, \ref{fig:main-acc}, \ref{fig:langs-acc} and \ref{fig:cptft-acc}.

\begin{figure}[ht]
\centering
\includegraphics[width=0.35\textwidth]{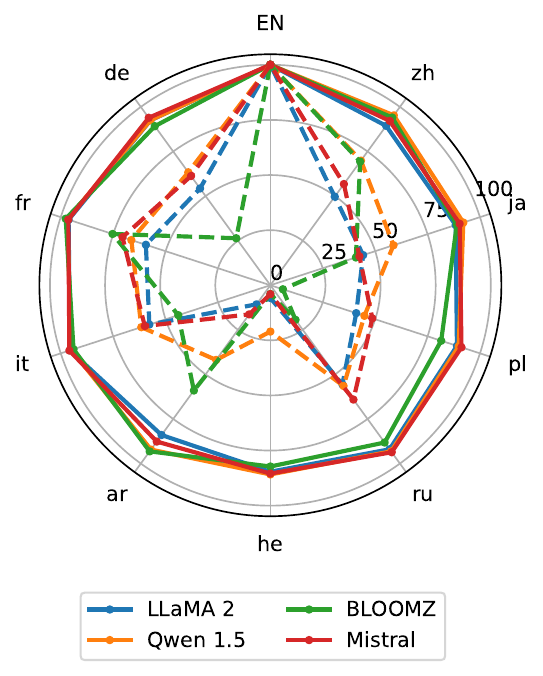}
\caption{XLTR of different models on QASC. Solid lines: WF-2 results; Dashed lines: NF results.}
\label{fig:qasc-xltr}
\end{figure}

\begin{figure}[ht]
\centering
\includegraphics[width=0.35\textwidth]{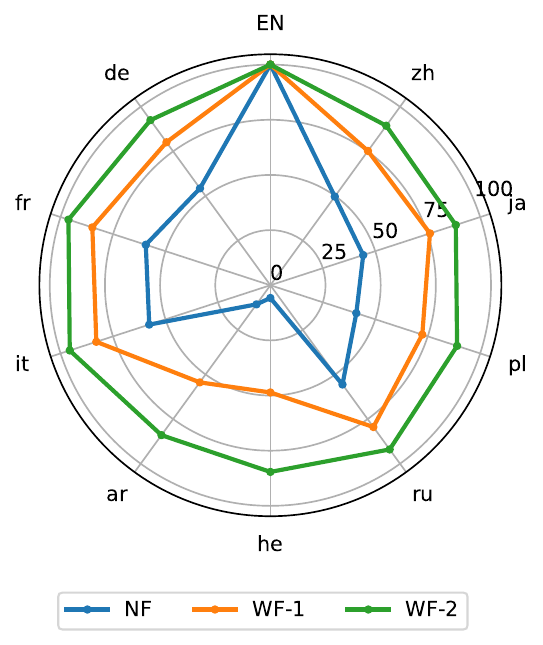}
\caption{XLTR of LLaMA-2-7B-Chat on QASC. Here WF-2 equals to WF-all, as QASC only has two pieces of evidence per sample.}
\label{fig:qasc-llama2-xltr}
\end{figure}

We provide the results of the QASC dataset in Figure~\ref{fig:qasc-xltr} and \ref{fig:qasc-llama2-xltr}.
The results show that the more knowledge provided leads to better cross-lingual transferability, which aligns with our conclusion.
% TODO: add description for the QASC dataset

\section{Evaluation on MMLU}
\label{sec:app-eval-mmlu}
In order to address the concerns of overfitting on the fine-tuned datasets and forgetting the world knowledge, we also evaluate the LLaMA-2-7B-Chat model on MMLU before and after fine-tuning.
The results are listed in Table~\ref{tab:finetuned-mmlu}.
We find that the model's MMLU performance shows both increases and decreases after finetuning, with relatively small changes in magnitude.
This indicates that we do not cause the model to overfit these datasets and maintain its general capabilities.

\begin{table}
\centering
\footnotesize
\begin{tabular}{l|c}
\hline
\textbf{Finetuned Dataset} & \textbf{MMLU(5-shot)} \\
\hline
None & 40.88 \\
KFRD Arithmetic & 45.53 \\
KFRD Symbolic & 43.87 \\
KFRD Logical & 45.14 \\
StrategyQA WF-all & 46.85 \\
ASDiv & 38.31 \\
Coin Flip & 43.58 \\
ProofWriter & 37.72 \\
\hline
\end{tabular}
\caption{The accuracy of LLaMA-2-7B-Chat model on MMLU both before and after finetuning}
\label{tab:finetuned-mmlu}
\end{table}

\section{Language Distribution of Model Training Corpora}
\label{sec:app-lang-dist}
% In this section, we present the language distribution of the pre-training corpora, referencing Table \ref{tab:llamalanguage-distribution} from the LLaMA2 paper~\citep{touvron2023llama} and Table~\ref{tab:bloomzlanguage-distribution} from the BLOOM paper~\citep{workshop2022bloom}.
In this section, we discuss the language distribution of the pre-training corpora, from the LLaMA2 paper~\citep{touvron2023llama} and the BLOOM paper~\citep{workshop2022bloom}.
Unfortunately, we were unable to locate the corresponding distribution data for Mistral and Qwen. 

For LLaMA2, languages such as Arabic and Hebrew were not included in the table provided in \citet{touvron2023llama} (Table~10), indicating that their proportions are lower than 0.005\%, categorizing them as extremely low-resource languages.
The other eight languages discussed in the paper are represented.
Notably, German and Chinese rank as high-resource languages, accounting for 0.17\% and 0.13\% of the corpus, respectively, holding the second and fifth highest positions.

For BLOOM, only English, French, Chinese, and Arabic are explicitly listed in the paper (\citealp{workshop2022bloom}, Table~1), while other languages are not reported in the paper, indicating they are low-resource languages.

\section{Reasons for Creating a New Dataset}
\label{sec:app-why-create}
The primary reason for creating a new dataset is that most existing datasets involve knowledge retrieval, which does not align with our focus on knowledge-free reasoning. For instance, in StrategyQA, while necessary reasoning knowledge is provided, it may be incomplete.

\textbf{StrategyQA Example:}
\begin{itemize}
    \item \textbf{Question:} Are you likely to find a crucifix in Karachi?
    \item \textbf{Facts:} The crucifix is a symbol of Christianity. The vast majority of Pakistan's population is Muslim.
    \item \textbf{Missing Knowledge:} It is not specified that Karachi is in Pakistan.
\end{itemize}

Similarly, most existing math datasets also require knowledge retrieval to answer questions, such as the ASDiv-a dataset.

\textbf{ASDiv-a Example 1:}
\begin{itemize}
    \item \textbf{Question:} At the school's book fair, Sam bought 13 adventure books and 17 mystery books. If 15 of the books were used, how many new books did he buy?
    \item \textbf{Missing Knowledge:} The new books are those that were not used.
\end{itemize}

\textbf{ASDiv-a Example 2:}
\begin{itemize}
    \item \textbf{EN-Question:} After the aviary was the zoo's swamp area. Penny counted a total of 55 tree frogs, 10 poison frogs, and 13 wood frogs. How many frogs was Penny able to count?
    \item \textbf{FR-Question:} Après la volière se trouvait la zone marécageuse du zoo. Penny a dénombré un total de 55 rainettes, 10 grenouilles venimeuses et 13 grenouilles des bois. Combien de grenouilles Penny était-elle capable de compter ?
    \item \textbf{Missing Knowledge:} In English, it can be inferred that ``poison frogs," ``wood frogs," and ``tree frogs" are all ``frogs." However, in French, it is not directly inferable that ``rainettes" are a type of ``grenouilles," requiring additional knowledge retrieval.
\end{itemize}

% 而一些现存的逻辑数据集，其不是为了knowledge-free设计，使用的是现存的真实实体。这导致了虽然可以理论上通过不检索知识去回答答案，但是检索知识可能会影响最终的答案。例如For example, "Harry is a cat." The model can infer that "Harry is an animal" based on existing knowledge without requiring contextual reasoning rules. 基于现存的知识，模型也更可能认为“The squirrel likes the squirrel.” 这会影响最终的结果，特别是当结果是不相关的时候。模型可能会依据现存知识，去判断出相关。当叠加翻译的问题时候，这个问题会更加显著。例如当EN->ZH时，“The squirrel likes the squirrel.”将被翻译成“squirrels likes squirrels.”. 因为在中文中，不存在冠词。这会导致现存知识的影响更大。进而导致做错题目。 当我们在训练集和测试集中去掉不相关的问题时，结果如图15所示。Proofwrite的传递率会显著的上升，near fully cross-lingual transfer。这进一步证明了这一点。

Some existing logic datasets are not designed with knowledge-free reasoning in mind, as they use real-world entities. This leads to situations where, although it is theoretically possible to answer without retrieving external knowledge, the retrieval of such knowledge might influence the final answer. For example, given the statement ``Harry is a cat," the model might infer ``Harry is an animal" based on its existing knowledge, without requiring contextual reasoning rules. Similarly, based on prior knowledge, the model might incorrectly assume ``The squirrel likes the squirrel" as related, especially when the actual context is irrelevant. 

This issue becomes more pronounced when translation is involved. For instance, when translating from English to Chinese, ``The squirrel likes the squirrel" may become ``squirrels like squirrels," as Chinese does not use articles. This can amplify the influence of pre-existing knowledge, leading to incorrect answers.
% When we removed irrelevant questions from the training and testing sets, as shown in Figure \ref{xxx}, the transfer ratio for ProofWriter significantly improved, achieving near-full cross-lingual transfer. This further supports our hypothesis.

% \textbf{Translation Issues:}
By constructing our own dataset, we also avoid potential translation issues that arise when existing datasets are used in different languages, ensuring that reasoning tasks are uniformly understood across languages.

Another advantage of creating a new dataset is that we can control the difficulty level. If the dataset is too difficult and models have low accuracy in English, it would be meaningless to measure cross-lingual transferability. Moreover, a new dataset allows for a more comprehensive coverage of reasoning operations.

% \begin{figure}[ht]
% \centering
% \includegraphics[width=0.35\textwidth]{fig/sqa-acc.pdf}
% \caption{Accuracy of different models on StrategyQA. Solid and dashed line represent the result of With Facts and No Facts setting, respectively.}
% \label{fig:sqa-acc}
% \end{figure}

% \begin{figure}[ht]
% \centering
% \includegraphics[width=0.35\textwidth]{fig/sqa-llama2-acc.pdf}
% \caption{Accuracy of LLaMA 2 7B Chat on StrategyQA under various settings.}
% \label{fig:sqa-llama2-acc}
% \end{figure}

% \begin{table}
% \centering
% \begin{tabular}{l|cc}
% \hline
% \textbf{Dataset} & \textbf{Samples} & \textbf{Epoch}\\
% \hline
% Fiction Knowledge & 2000 & 4 \\
% MKQA & 5758 & 1 \\
% KFRD Arithmetic & 8000 & 1\\
% KFRD Symbolic & 10000 & 1\\
% KFRD Logical & 10000 & 1\\
% \hline
% \end{tabular}
% \caption{Training epoch and number of samples of fine-tuned datasets in the intepretation experiments}
% \label{tab:data-inte-epoch-samples}
% \end{table}

%%% 需要进行修改位置
\input{tab/dataSetExamples}

% \section{Additional Works on Cross-lingual Transfer}
% \label{sec:addWorks}

% \citet{devlin2018bert} introduced mBERT, advancing cross-lingual transfer by capturing shared linguistic patterns in a unified embedding space, enabling zero-shot transfer without parallel corpora.
% Similarly, \citet{conneau2019unsupervised} showed XLM's effectiveness in optimizing multilingual embeddings, improving performance in translation and classification tasks.
% \citet{ansell2021composable} proposed composable sparse fine-tuning, selectively fine-tuning sparse parameters across languages to reduce interference and boost performance, especially in low-resource settings, outperforming adapter-based methods in tasks like NER and NLI.

% Additionally, \citet{ye2023language} assessed the imbalance of knowledge across different languages in LLMs, observing weak cross-lingual transferability of knowledge. \citet{zhu2024question} discovered that training on translated questions can enhance the cross-lingual transferability of reasoning tasks. 

\end{document}

%% file: tab/exampleKFRD.tex
\begin{table*}[ht]
\centering
\footnotesize
\renewcommand{\arraystretch}{1}
\setlength{\tabcolsep}{4pt} % 缩小列之间的间距
\scalebox{0.85}{
\begin{tabular}{l|m{12cm}} % 行宽，越宽可以越少行
\hline
\multicolumn{2}{c}{\textbf{Arithmetic Reasoning}} \\
\hline
\textbf{Input} &
11,  645 (two numbers)
\\
\hline
\textbf{Transformation Rule} &
Addition (a mathematical operation)
\\
\hline
\textbf{Output Options} &
\makecell[l]{A) 595 \quad B) 536 \\ C) 771 \quad \textbf{D) 656}}
\\
\hline
\multicolumn{2}{c}{\textbf{Symbolic Reasoning}} \\
\hline
\textbf{Input} &
education, game, president, night, man (3-5 words in the corresponding language)
\\
\hline
\textbf{Transformation Rule} &
Swap the positions of the 5th and 2nd words; Delete the 2nd word (1-3 symbolic operations)
\\
\hline
\textbf{Output Options} &
\makecell[l]{\textbf{A) education, president, night, game} \quad B) education, problem, night, game \\ C) hand, president, night, game \quad D) education, house, night, game}
\\
\hline
\multicolumn{2}{c}{\textbf{Logical Reasoning}} \\
\hline
\textbf{Input} &
Alex is Aurora Vale. Everything that is Aurora Vale is Omicron Delta. Stella is not Chronos Wasteland. Max is not Dreamweaver's Haven. Suppose Sally is Whispering Meadows, then Sally is Chimerical Citadel. Everything that is Ebonwyrm Abyss is Phoenixfire Ridge. (6 propositions)
\\
\hline
\textbf{Transformation Rule} &
Implication Elimination (a logical rule)
\\
\hline
\textbf{Output Options} &
\makecell[l]{A) Alex is Seraphim Heights. \quad B) Alex is Tempestwilds. \\ \textbf{C) Alex is Omicron Delta.} \quad D) Polly is Arcadia Reach.}
\\
\hline
\end{tabular}
}
\caption{Examples of different tasks in the KFRD dataset}
\label{tab:exampleKFRD}
\vspace{-0.2cm}
\end{table*}

%% file: tab/sqaExample.tex
\begin{table*}[ht]
\centering
\footnotesize
\renewcommand{\arraystretch}{1.1}
\begin{tabular}{l|m{11cm}}
\hline
\multicolumn{2}{c}{\textbf{StrategyQA}} \\
\hline
\textbf{Question} & Are more people today related to Genghis Khan than Julius Caesar? \\
\hline
\textbf{Facts} &
\makecell[l]{
1. Julius Caesar had three children. \\
2. Genghis Khan had sixteen children. \\
3. Modern geneticists have determined that out of every 200 men today has DNA \\ \ \ \ \ that can be traced to Genghis Khan.
} \\
\hline
\textbf{Answer} & Yes \\
\hline
\multicolumn{2}{c}{\textbf{QASC}} \\
\hline
\textbf{Question} & Climate is generally described in terms of what? \\
\hline
\textbf{Facts} &
\makecell[l]{
1. Climate is generally described in terms of temperature and moisture. \\
2. Fire behavior is driven by local weather conditions such as winds, temperature and \\ \ \ \ \ moisture. 
} \\
\hline
\textbf{Options} &
\makecell[l]{
A. sand \\
B. occurs over a wide range \\
C. forests \\
D. Global warming \\
E. rapid changes occur \\
F. local weather conditions \\
G. measure of motion \\
H. city life
} \\
\hline
\textbf{Answer} & F \\
\hline
\end{tabular}
\caption{Examples of knowledge-involved datasets}
\label{tab:sqaExample}
\end{table*}

%% file: tab/dataSetExamples.tex
\begin{table*}[ht]
\centering
\footnotesize
\renewcommand{\arraystretch}{1.1}
\begin{tabular}{l|m{11cm}}
\hline
% \multicolumn{2}{c}{\textbf{StrategyQA}} \\
% \hline
% \textbf{Question} & Are more people today related to Genghis Khan than Julius Caesar? \\
% \hline
% \textbf{Facts} &
% \makecell[l]{
% 1. Julius Caesar had three children. \\
% 2. Genghis Khan had sixteen children. \\
% 3. Modern geneticists have determined that out of every 200 men today has DNA \\ \ \ \ \ that can be traced to Genghis Khan.
% } \\
% \hline
% \textbf{Answer} & True \\
% \hline
\multicolumn{2}{c}{\textbf{MKQA}} \\
\hline
\textbf{Query} & Who sings "I Hear You Knocking But You Can't Come In"? \\
\hline
\textbf{Answers} & Dave Edmunds \\
\hline
\multicolumn{2}{c}{\textbf{BoolQ}} \\
\hline
\textbf{Question} & Do Iran and Afghanistan speak the same language? \\
\hline
\textbf{Answer} & True \\
\hline
\multicolumn{2}{c}{\textbf{AmbigQA}} \\
\hline
\textbf{Question} & How often does spermatogenesis—the production of sperm—occur? \\
\hline
\textbf{Answer} & 74 days \\
\hline
\end{tabular}
\caption{Examples of adapted datasets used in this paper}
\label{tab:dataSetExamples}
\end{table*}